\documentclass[10pt,journal,compsoc]{IEEEtran}
\newif\ifpeerreview

\peerreviewfalse

\usepackage[nocompress]{cite}
\usepackage{url}
\usepackage{amsmath,amssymb,graphicx}

\usepackage{lipsum} % Only used to generate random text.

\usepackage{graphicx}
\usepackage{amsmath}
\usepackage{amssymb}
\usepackage{booktabs}
\usepackage{float}
\usepackage{stfloats}
\usepackage{tikz}
\usepackage[pagebackref=true,breaklinks=true,colorlinks,bookmarks=false]{hyperref}
\usetikzlibrary{spy,backgrounds}
\usepackage{xcolor}
\usepackage{mdframed}

\usepackage[switch]{lineno}

\newcommand{\paperID}{15}

\newcommand{\highlight}[1]{{#1}}    % use this command to generate normal output
\title{Aberration-Aware Depth-from-Focus}

\author{Xinge~Yang,~\IEEEmembership{Student~Member,~IEEE,}
        and~Qiang~Fu,
        and~Mohamed~Elhoseiny,~\IEEEmembership{Member,~IEEE,}
        and~Wolfgang~Heidrich,~\IEEEmembership{Fellow,~IEEE}% <-this % stops a space
\IEEEcompsocitemizethanks{\IEEEcompsocthanksitem W. Heidrich is with the Department
of Computer Science and Electrical and Computer Engineering, King Abdullah University of Science and Technology, Saudi Arabia, 23955.\protect\\
E-mail: wolfgang.heidrich@kaust.edu.sa
\IEEEcompsocthanksitem X. Yang, Q. Fu, M. Elhoseiny is with King Abdullah University of Science and Technology.}% <-this % stops an unwanted space
}

\begin{document}

\IEEEtitleabstractindextext{%
\begin{abstract}
Computer vision methods for depth estimation usually use simple camera models with idealized optics. For modern machine learning approaches, this creates an issue when attempting to train deep networks with simulated data, especially for focus-sensitive tasks like Depth-from-Focus. In this work, we investigate the domain gap caused by off-axis aberrations that will affect the decision of the best-focused frame in a focal stack. We then explore bridging this domain gap through aberration-aware training (AAT). Our approach involves a lightweight network that models lens aberrations at different positions and focus distances, which is then integrated into the conventional network training pipeline. We evaluate the generality of network models on both synthetic and real-world data. The experimental results demonstrate that the proposed AAT scheme can improve depth estimation accuracy without fine-tuning the model for different datasets. The code will be available in \href{https://github.com/vccimaging/Aberration-Aware-Depth-from-Focus}{github.com/vccimaging/Aberration-Aware-Depth-from-Focus}.
\end{abstract}

}

% Make Title
\ifpeerreview
\linenumbers \linenumbersep 15pt\relax 
\author{Paper ID \paperID\IEEEcompsocitemizethanks{\IEEEcompsocthanksitem This paper is under review for ICCP 2023 and the PAMI special issue on computational photography. Do not distribute.}}
\markboth{Anonymous ICCP 2023 submission ID \paperID}%
{}
\fi
\maketitle
\thispagestyle{empty}

% The first section title should be wrapped inside a \IEEEraisesectionheading as follows.
% \IEEEraisesectionheading{
%   \section{Introduction}\label{sec:introduction}
% }
% The very first letter of the paper is a 2 line initial drop letter
% followed by the rest of the first word in caps.
% \IEEEPARstart{T}{his} demo file is intended to serve as a ``starter
% file'' for ICCP 2023 submissions produced under
% \LaTeX~\cite{kopka-latex} using IEEEtran.cls version 1.8b and later.

% For submissions for review, the paper uses the lineno package which
% might require you to compile under latex twice to get the line numbers
% to align correctly.

% \section{Introduction}

\IEEEraisesectionheading{
  \section{Introduction}\label{sec:introduction}
}

% DL with pinhole camera/thin lens model is inaccurate and has domain-gap because real camera lenses have both dof and aberration. 
\IEEEPARstart{M}{odern} deep-learning techniques have made significant progress in understanding 3D scenes from 2D RGB images, including depth estimation~\cite{niklaus20193d, bhat2021adabins, xie2022revealing, patil2022p3depth}, object detection~\cite{redmon2016you,zhang2022dino}, multiple views~\cite{huang2018deepmvs,ummenhofer2017demon}, and camera tracking and mapping~\cite{bloesch2018codeslam,tang2018ba,tateno2017cnn}. However, these methods assume that 2D training images are aberration-free, relying on an idealized pinhole camera model that fails to account for out-of-focus effects and optical aberrations present in real camera lenses. This inaccuracy leads to a domain gap between experimental and real-world images, particularly given that modern lenses possess large apertures resulting in shallow depth-of-field (DoF), as well as a large field-of-view resulting in off-axis aberrations. The domain gap undermines the generalizability of trained deep learning models~\cite{facil2019cam}, necessitating engineers to fine-tune models with real data for each end device. 

% existing works on focal stack DfD, interpolate sharp areas. they propose this is domain-invariant, but they ignore the optical domain gap
% moreover, this optical aberration is observed to affect the depth estimation
Recent studies in depth-from-focus (DfF)~\cite{suwajanakorn2015depth,
  anwar2017depth, carvalho2018deep, hazirbas2018deep,
  srinivasan2018aperture, gur2019single, maximov2020focus,
  wang2021bridging, won2022learning, yang2022deep} have recognized the
impact of out-of-focus effects and explored using defocus cues for
depth estimation and all-in-focus image estimation. DfF methods
estimate the probability that a pixel is the sharpest in a focal stack
and then interpolate the input focus distances based on the probability, assuming that
the sharpest frame is the best-focused frame. The probability can also be used to interpolate focused images for all-in-focus image synthesis~\cite{zhang2021deep, wang2021bridging, broad2016light,ruan2021aifnet}. Although some of these
methods~\cite{gur2019single, maximov2020focus, wang2021bridging,
  won2022learning, yang2022deep} have claimed to bridge the domain gap
between experimental and real-world images by considering that the sharpest pixel in the focal stack is independent of semantic
information, their experiments have relied on a thin lens model that
neglects off-axis optical aberrations commonly present in real
lenses. Such aberrations, including field curvature, can cause the
focus distance to vary across the image plane or result in asymmetric
blurs that make it difficult to measure the most in-focus accurately
frame.  Moreover, studies~\cite{trouve2013passive, chang2019deep,
  baek2021single, marquez2021snapshot} have highlighted the impact of
optical aberrations in depth estimation. Therefore, another domain gap
arises due to the presence of optical aberrations.

% aberration aware training solves the problem
To overcome the domain gap resulting from optical aberrations, we
introduce aberration-aware training (AAT) that enables the network to
learn these optical aberrations during the training. Our AAT method
consists of a lightweight point spread function (PSF) network and a
re-rendering process to simulate aberrated training images. First, we
compute the spatially-varying PSF of an optical lens using ray
tracing~\cite{wang2022differentiable,yang2023curriculum}, and train a
multilayer perceptron (MLP) to represent it. Once trained, the network
can efficiently estimate the PSF for different object positions and
focus distances. Next, we render training images to apply off-axis
aberrations. Given the depth map for each image, we select a group of
focus distances and determine the PSF for each pixel. Then we perform
local convolution with all-in-focus images to create a focal stack
containing both depth-of-field and off-axis aberrations. During the
training of the depth estimation network, the network learns to
determine the best-focused frame for each pixel under optical
aberrations, thereby enhancing the generalizability. Furthermore, the
depth estimation accuracy can be improved as long as the training and
testing images are captured/simulated using the same lens.

% experiemental evaluation. we find the network trained and evaluated with different lens models cannot generalize well. but with the same lens model the results are much better
To assess the effectiveness of our proposed AAT scheme, we conduct
experiments to test the generalizability of the pre-trained DfF model on
various datasets, including both simulated and real-world
datasets. First, we simulate focal stacks using a real lens and a
thin lens and train the corresponding DfF models. Then we test two
models on focal stacks simulated/captured by the same lens without any
fine-tuning. The experimental results demonstrate that the AAT
model generalizes better than the non-AAT model in terms of depth
estimation accuracy. Specifically, the AAT model successfully resolves
small depth differences between adjacent objects, whereas the non-AAT
model fails to do so. Moreover, the AAT model proves more successful
at suppressing the transfer of color texture detail into the depth
geometry. We demonstrate results on two recent state-of-the-art architectures~\cite{yang2022deep, wang2021bridging}, and the results show that our training approach is agnostic to the specific DfF network architecture.

%We evaluate two different
%DfF network architectures, and the results indicate that the AAT
%scheme can extract more hidden information under optical aberrations
%and improve both networks' generalizability. Overall, our experimental
%results demonstrate the effectiveness of the AAT scheme in improving
%the generalizability of deep learning models across different domains.

\highlight{In summary, our contributions are two-fold:

\begin{itemize}
    \item We propose a lightweight network that can represent the PSF of a real lens at different focus distances and object positions. This PSF network can then simulate aberrated and realistic images for aberration-aware training.
    \item We reveal the domain gap between real-world and simulated data arising from off-axis aberrations, which impacts the determination of the best-focused frame in a focus stack and reduces the accuracy of depth estimation. We propose an AAT scheme to address the problem, the DfF models trained with the AAT scheme can better generalize to test focal stacks captured or simulated by the same lens.
\end{itemize}
}
\section{Related Works}

\subsection{Depth from Focus}

Depth estimation is a fundamental task in computer vision, and can be tackled using different cues, such as semantic information~\cite{bhat2021adabins, patil2022p3depth}, stereo~\cite{godard2017unsupervised, chang2018pyramid, shen2021cfnet}, and defocus~\cite{wang2021bridging, yang2022deep, maximov2020focus}. Among these cues, the defocus cue is considered domain invariant, as the defocus pattern is only related to the optical properties of the imaging lens. The depth-from-focus problem learns the depth map from the focus stack, using the idea that each pixel must have one frame that is best in focus.

% methods, network architectures
Conventional approaches solve the depth-from-focus (DfF) problem as an optimization task~\cite{suwajanakorn2015depth, surh2017noise}. Learning-based approaches then use 2D convolutional neural networks (CNNs) to improve the accuracy of feature extraction and depth estimation~\cite{hazirbas2018deep, carvalho2018deep, gur2019single, maximov2020focus}. To allow for communication between different 2D CNNs, a shared pooling layer is used, and an intermediate defocus map~\cite{maximov2020focus} is learned during training to aid the depth estimation. Later, 3D CNNs~\cite{wang2021bridging, won2022learning} started to replace 2D architectures, as they perform better in extracting shared information from an image stack. While the idea of the intermediate defocus map is kept, it is replaced by a 3D cost volume~\cite{wang2021bridging, won2022learning, yang2022deep}.

% % datasets
The training and testing datasets are obtained through various methods, including real-world captures and image simulation. For real-world captures, focal swap~\cite{suwajanakorn2015depth, carvalho2018deep} and light-field camera rendering~\cite{hazirbas2018deep} are two common methods used to obtain focused images, and additional Lidar or ToF cameras are used to get the depth map of the scene. Capturing real-world focal stacks with ground truth depth is quite expensive. Thus, image simulation methods, including PSF convolution~\cite{gur2019single}, stereo image rendering~\cite{wang2021bridging}, and software rendering~\cite{maximov2020focus}, are also used to generate simulated focal stacks for training data augmentation. However, existing simulated focal stacks all rely on the idealized thin lens model and thus lack optical aberrations. Moreover, the real-world captured focal stacks lack focus distance information and lens data. In this work, we simulate aberrated focal stacks with the real lens model and capture real-world focal stacks with the necessary information. We use simulated focal stacks for training, and after training, we test the models on both simulated and real-world focal stacks without fine-tuning.

\subsection{PSF Estimation}

Multiple approaches have been introduced to calculate the PSF of an optical lens/system. The commonly used idealized optical model~\cite{freeman1990optics, gur2019single} calculates the PSF under paraxial principles, resulting in a truncated Gaussian function called the ``Circle-of-Confusion'' (CoC). The diameter of the CoC is computed by

\begin{equation}
    CoC=\frac{f}{N} \frac{\left|z - f_{d}\right|}{z} \frac{f}{f_{d} - f},
    \label{equ:coc}
\end{equation}
where $f_{d}$ is the focal distance, $z$ is the distance between an object to the lens, $f$ is the focal length, and $N$ is the F-number of the camera lens. However, due to the paraxial approximation, this approach can not represent spatially varying PSF and off-axis aberrations like field curvature, coma, and astigmatism in actual optical lenses.

Ray tracing through optical lenses~\cite{kolb1995realistic, wang2022differentiable, sun2021end} can provide a more accurate spatially-varying PSF and has been widely utilized in commercial software such as ZEMAX and CodeV. This approach involves shooting rays from each object point source and calculating the distribution of each ray on the sensor plane to obtain the PSF. However, tracing rays through a sequence of optical elements is computationally expensive as it requires finding the intersection points of each ray using iterative algorithms. Recently, researchers have started exploring the use of neural networks to represent the PSF~\cite{tseng2021differentiable}, which has shown promising results in both accuracy and speed. However, nobody has explored representing a lens with both varying focus distances and spatial positions. In this work, we extend the network estimation approach to represent the PSF for varying focus distances and spatial positions.

\highlight{Another widely used method, especially when the design space of optical lenses is unavailable, is PSF calibration~\cite{li2019depth, chen2020data}. PSF calibration approaches measure the lens response to the given point light source and then estimate the PSF at unmeasured positions~\cite{rego2021robust, yanny2020miniscope3d, denis2015fast, denis2011fast, sroubek2016decomposition}. Among all the PSF estimation works, the recently proposed low-rank model~\cite{yanny2020miniscope3d, yanny2022deep} is an efficient and accurate way to estimate unmeasured PSF with only a few measurements. However, PSF calibration also comes with its own challenges, such as noise and quantization (especially for long tails), since PSF estimation requires either the direct measurement of (weak) point sources or the solution of a deconvolution problem.}

% \textcolor{red}{introduce PSF calibration and low-rank model}
\section{Methodology}

\begin{figure*}[ht]
    \centering
    \includegraphics[width=\textwidth]{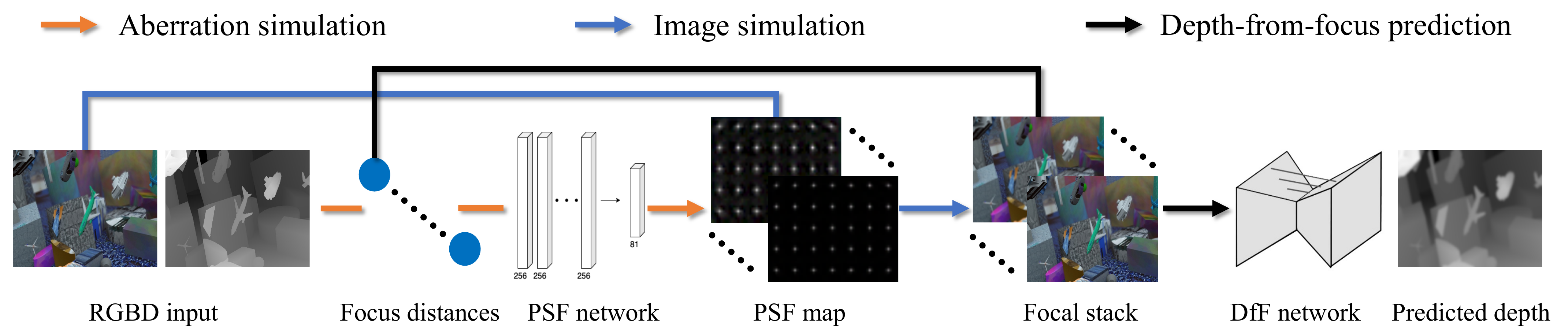}
    \caption{\textbf{Aberration-Aware Training pipeline.} (a) all-in-focus RGB images and corresponding depth maps are given as the input. (b) different focus distances are selected to simulate the focal swap process. The PSF network estimates PSF for different object positions and focus distances (orange path). Then the PSF is convolved with the all-in-focus image to get the focal stack (blue path). (c) the DfF network takes the focal stack and focus distances to estimate the depth map (black path).}
    \label{fig:pipeline}
\end{figure*}

As shown in Fig.~\ref{fig:pipeline}(c), a classical DfF network takes the focus stack and corresponding focus distances as input and outputs a depth map. However, in the existing training pipeline, the focal stacks are pre-given and the network is blind to the optical system during training. Therefore, the pre-trained model is hard to generalize to real-world data, and computer vision engineers are required to fine-tune the model for different end devices. With the idea of embedding optical characteristics into the network training, the AAT scheme integrates the data simulation module into the network training pipeline, as illustrated in Fig.~\ref{fig:pipeline}(b). All-in-focus images and their corresponding depth maps are used as input during the training process (Fig.~\ref{fig:pipeline}(a)). A series of focus distances are selected, and the PSF estimation network is applied to estimate the PSF for each pixel, which is then used to render the focal stack. The entire pipeline can be executed end-to-end, and the AAT scheme allows the network to learn the estimation of depth maps in the presence of complex optical aberrations.

\subsection{Depth-from-Focus Network}

We do not modify the architecture of the DfF network but instead use existing architectures such as AiFNet~\cite{wang2021bridging} and DFVNet~\cite{yang2022deep}. Here we use AiFNet for illustration. Given a batch of focal stacks with shapes (B, S, C, H, W), where B is the batch size, S is the stack size, C is the number of channels, and H, W are the image height and width, a 3D convolutional encoder is used to extract multilayer image features and create an intermediate attention map. This attention map functions as an in-focus probability map, which is then used to interpolate focus distances for the final depth estimation. The loss function for the DfF task typically consists of a depth estimation loss and a regularization term:

\begin{equation}
    L = L_{depth} + \omega L_{reg},
    \label{equ:loss}
\end{equation}
where $L_{depth}$ denotes the supervision loss function designed directly on the estimated depth map, and $L_{reg}$ denotes the regularization term (e.g., one that encourages the depth map to be locally smooth using an edge-aware weighting as in~\cite{godard2017unsupervised}). $\omega$ is the weight coefficient balancing the two parts, and we adopt the hyperparameter settings from the original paper.

% Due to the ideal thin lens model, existing DFF methods ignore optical aberrations, resulting in degraded performance when applying the pre-trained model to real-world focal stacks. However, the AAT scheme helps train the network to align with the optical lens, which allows the network to extract more accurate image features in the presence of optical aberrations. 

% \subsection{PSF network}
\subsection{Aberration Simulator}
% \WH{Changed the title to focus on the optical simulation part. For the
%   network name, maybe just ``PSF Network'' or ``Focus Network''?}

\begin{figure}[ht]
    \centering
    \includegraphics[width=0.98\linewidth]{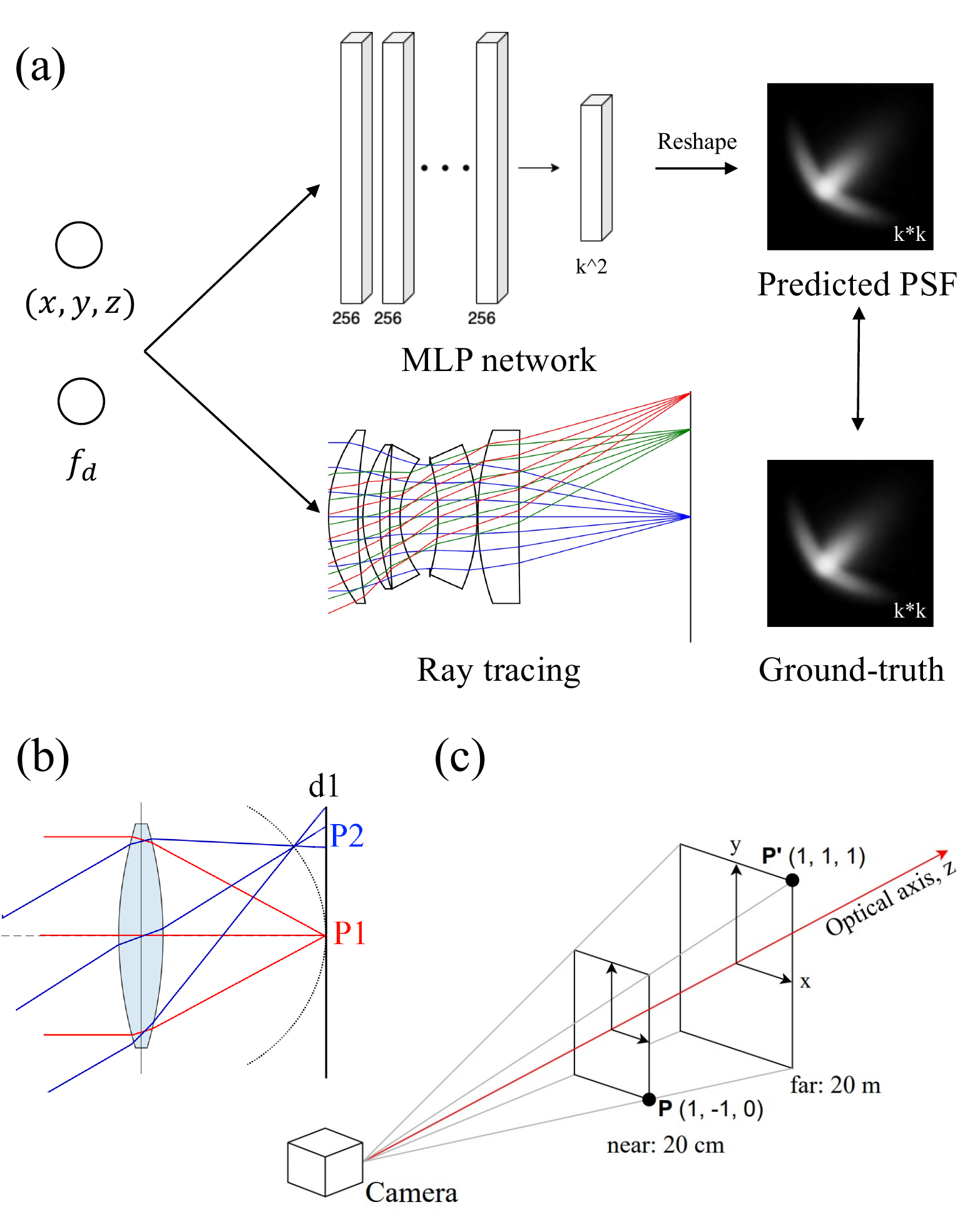}
    \caption{(a) a MLP network is trained to represent the PSF for different positions and focus distances. We use ray tracing to calculate accurate PSF as the ground truth. The network takes as input the object positions $(x,y,z)$ and focus distance $f_d$, and produces a 2D matrix as output. (b) off-axis optical aberrations blur the pixels at the edge (P2) when the pixel on the axis (P1) is focused. (c) the valid imaging area is a frustum, and we normalize the $x, y$ coordinates to [-1, 1], $z$ and $f_d$ to [0, 1].}
    \label{fig:mlp}
\end{figure}

Our contribution lies in the accurate simulation of aberrations for
the training data to improve on the classical thin lens model.  The
PSF characterizes the optical lens response to a point source of
light. We can convolve the per-pixel PSF with the object image to
simulate the image captured by a camera.
%A commonly used PSF model is the constrained Gaussian function
%(hereafter referred to as the "Gaussian PSF") that accounts for energy
%dispersion when the object is out-of-focus, as described by
%Eq.~\ref{equ:coc}.
The Gaussian PSF (Eq.~\ref{equ:coc}) assumes shift-invariance across the same
depth plane. However, this idealized optical model does not accurately account for
off-axis optical aberrations in a real camera lens. In
Fig.~\ref{fig:mlp}(b), we show the field curvature aberration, which
is common in real lenses. Due to the field curvature, the pixel at the
edge (P1) is blurry, but the pixel on the axis (P2) is sharp, although this image is well-focused to depth $d_1$.

Ray tracing through optical lenses is a well-established technique for obtaining a more accurate point spread function (PSF)~\cite{kolb1995realistic, wang2022differentiable}. This involves tracing a group of rays from a point source through the lens group to the sensor plane, resulting in a spot diagram. We can then convert the spot diagram into sensor pixels and obtain the PSF~\cite{freeman1990optics,jenkins1957fundamentals,li2021end} by:

% ====================================>

% \begin{equation}
%     \mathrm{PSF} = \sum_{k=1}^{spp} \mathbf{o}_{k} \cdot \delta_{i, j}(\mathbf{o}_{k}),
% \end{equation}
% where $\delta_{i,j}$ denotes the energy distribution of a ray at
% pixel $(i, j)$. The term $spp$, short for ``samples per pixel'', represents
% the number of rays emitted from each point source. In our experiments, we assume that the energy
% distribution of a ray is a unit impulse that equals zero except within
% the pixel. 

% ====================================>
\begin{equation}
    \begin{aligned}
       \mathrm{PSF}(\mathbf{o_p}) = \sum_{k=1}^{spp} &u_{k} \cdot \sigma(|(\mathbf{o_p} - \mathbf{o_k}) \cdot \hat{\mathbf{e}}_x|/L) \\ 
       & \cdot \sigma(|(\mathbf{o_p} - \mathbf{o_k}) \cdot \hat{\mathbf{e}}_y|/L), 
    \end{aligned}
\end{equation}
where $\mathbf{o}_p$ denotes the coordinate of the pixel, and $\mathbf{o}_k$ represents the intersection point of the $k$th ray with the sensor plane. The variable $spp$ stands for ``samples per pixel'', corresponding to the number of rays emitted from each point source, which is set to 2048 in our experiments. We assume that the energy of each ray, denoted by $u_k$, is equal to 1. $\hat{\mathbf{e}}_x$ and $\hat{\mathbf{e}}_y$ are unit vectors in the sensor plane, and $L$ denotes the physical width of a sensor pixel. The $\sigma$ function is defined as:

\begin{equation}
    \sigma(x) = 
    \begin{cases}
        1-x & 0 \leq x \leq 1 \\
        0 & otherwise
    \end{cases},
\end{equation}
which assesses a ray's impact on its surrounding pixels, with a greater impact attributed to rays in closer proximity. The total impact of a ray on the four surrounding pixels sums to one. By leveraging sub-pixel information, the $\sigma$ function can more accurately represent the actual light distribution using a limited number of samples. Since DfF is based on imaging with large apertures, it is possible to neglect diffraction effects which would dominate in optical systems with a small aperture. Moreover, we assume that chromatic aberrations are well corrected compared to the other aberrations and out-of-focus effects, which is the case for most commercial-grade lenses. This allows us to simulate the optical system at a single wavelength of 589nm. 

However, ray tracing is computationally expensive, particularly as objects may appear at different positions and the focus distance may vary. Inspired by~\cite{tseng2021differentiable}, we train a network $H$ with parameters $\theta$ to represent the PSF as 

\begin{equation}
    % {\rm PSF}(x, y, z| f_{d}) = H(x, y, z, f_{d})
    \theta =\arg\min_{\theta} \|{\rm PSF}(x, y, z; f_{d}) - H_{\theta}(x, y, z; f_{d})\|_2^2,
    \label{equ:psf}
\end{equation}
where $(x,y,z)$ represent the normalized coordinates of an object point, and $f_d$ denotes the focus distance. As depicted in Fig.~\ref{fig:mlp}(a), we train the network $H$ to fit an imaging lens by minimizing the difference between the estimated PSF and the ray-traced PSF. In object space, the valid imaging region of a lens is a frustum (as shown in Fig.~\ref{fig:mlp}(c)) that is defined by the field of view (FoV), sensor size, minimum depth, and maximum depth. We then normalize the focus distance $f_{d}$ and depth $z$ to [0, 1], and normalize $(x, y)$ to [-1, 1]. 

Once the optical structure and aperture size of a lens is fixed, the PSF is solely determined by the object position and focus distance. To map the four-parameter input into a 2D PSF kernel, we adopt a simple MLP network. The PSF estimation network consists of one input layer, five hidden layers with 256 neurons each, and an output layer with $k^2$ neurons, where $k$ is the width of PSF. We use ReLU activation functions after each input and hidden layer and a Sigmoid activation function after the output layer. The $k^2$-channel output is then reshaped into a $k \times k$ 2D tensor. After training, we fix the parameters of the PSF network and use it to estimate the PSF for various object positions and focus distances. Then, we can use the per-pixel PSF to render aberrated images for the subsequent depth estimation task.

\section{PSF Estimation Results}
\label{sec:psf-rep}

\subsection{Implementation Details}

We train the PSF network for 400,000 iterations to overfit the lens. In
each iteration, we randomly focus the lens to a distance of $f_d$, and
uniformly select 256 points in object space for training. The ground-truth PSFs
are computed by tracing 1024 rays from each object point, and
$(x, y, z, f_d)$ coordinates are provided to the network as input. We
use a wavelength of 589~nm and set the PSF size to
${\rm 11}\times{\rm 11}$ sensor pixels. AdamW optimizer~\cite{loshchilov2017decoupled} and CosineAnnealing
learning rate scheduler~\cite{loshchilov2016sgdr} with default
parameters are used to train the PSF network. The initial learning rate is set to $1\times10^{-3}$. The
lens has a minimum imaging depth of 20cm and a maximum depth of 20m,
beyond which no relevant depth information can be recovered due to the
small baseline of the DfF approach. We use the same depth range for the
focus distance. We employ the DeepLens
framework~\cite{wang2022differentiable, yang2023curriculum} for ray
tracing computation. The training process is run on a single A100 80G GPU and
completed in approximately 6 hours. The network consists of 0.28
million parameters, with a storage size of approximately 1.5 MB.

\begin{figure*}[htp]
    \centering
    \includegraphics[width=\textwidth]{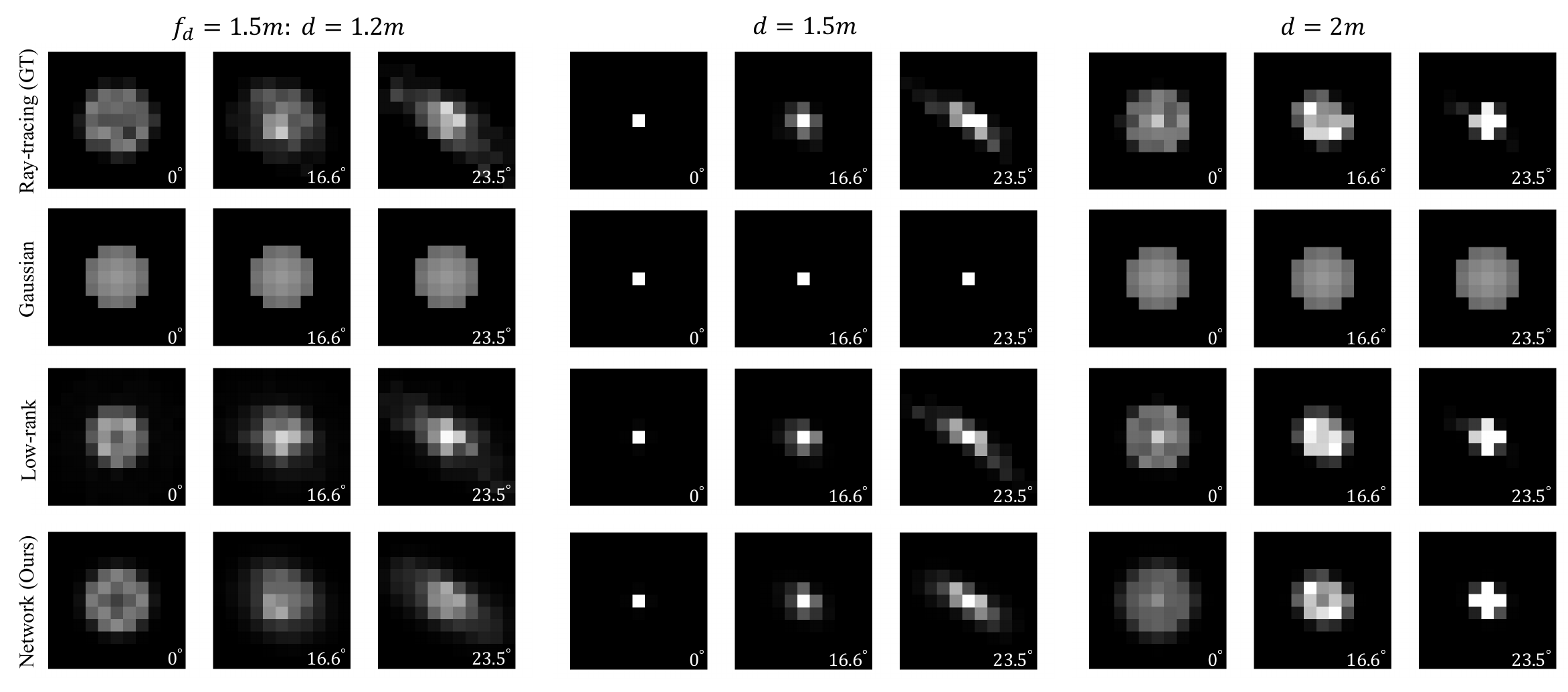}
    \caption{\textbf{PSF estimation results.} We focus the lens to a distance of 1.5m and evaluate the PSF of the four methods at three depths and three view angles. Both our PSF network and low-rank model produce a PSF that is close to the ground truth (ray tracing). In contrast, the Gaussian PSF exhibits significant differences, particularly at large view angles. The PSF of a real lens varies with different view angles due to the presence of off-axis optical aberrations, while the Gaussian PSF model neglects this aberration.}
    \label{fig:psf_pred}
\end{figure*}

\subsection{Evaluation}

% We simulate a 50mm F/2.8 lens generated by
% \href{http://lensnet.herokuapp.com/}{lensnet.com}~\cite{cote2021deep}
% for our experiments, as shown in Fig.~\ref{fig:mlp}. 

We use the Canon EOS RF50mm F/1.8 lens~\cite{US20210263286A1} for evaluation. The image sensor is simulated with physical dimensions of $24{\rm mm} \times 32{\rm mm}$ and a resolution of $640 \times 960$ to match the resolution of the RGBD data used for training the DfF network. While this image resolution is much lower than that of modern cameras, our experiments show that off-axis aberrations are still visible in the out-of-focus regime.

After training the PSF network, we focus the lens to a distance of
1.5m and evaluate the PSF at depths of 1.2m, 1.5m, and 2m, and three
different view angles, as shown in Fig.~\ref{fig:psf_pred}. At the
focused depth (1.5m), the PSF is small, but when the lens is out of
focus (1.2m and 2m), the PSF becomes larger. At $0^{\circ}$, both the
network and the Gaussian model accurately predict the ground truth
PSF. However, as the viewing angle increases, the Gaussian PSF becomes
increasingly inaccurate due to off-axis aberrations, and a clear
difference can be observed between the Gaussian PSF and the ground
truth at $23.5^{\circ}$. In contrast, the PSF network estimated
accurate results at all depths and view angles. Furthermore, at
$23.5^{\circ}$, the ground truth PSF at a depth of 2m is the smallest
among the three depths, caused by the field curvature  aberration
(also illustrated in Fig.~\ref{fig:mlp}(b)).
However, the Gaussian model can not capture this phenomenon. Comparing the network estimation and the ray tracing results, we find that the network-estimated PSF exhibits less noise and holds better symmetry, which is more in line with the physical situation. 
% To match this quality with direct ray tracing, more ray samples would be required for each PSF, which would drive up the simulation costs and is our motivation for injecting a neural network as a stand-in for the physical simulation.

\begin{table}[ht]
    \caption{Quantitative comparison of difference PSF estimation methods.}
    \centering
    \renewcommand{\arraystretch}{1.6}
    \resizebox{\linewidth}{!}{ 
    \begin{tabular}{l||c|c|c}
        \hline
        Method & $\ell_1$ error & $\ell_2$ error & time (min) \\
        \hline
        Ours & $\mathbf{4.68}\mathbf{e}^{\mathbf{-3}}$ & $\mathbf{8.23}\mathbf{e}^{\mathbf{-5}}$ & 2.5 \\
        \hline
        Tseng, et, al.~\cite{tseng2021differentiable} & $7.35e^{-3}$ & $2.94e^{-4}$ & \textbf{1.5} \\
        \hline
        Kyrollos, et, al.~\cite{yanny2022deep} & $7.33e^{-3}$ & $1.43e^{-3}$ & 87 \\
        \hline
    \end{tabular}
    }
    \label{tab:psf_comparison}
\end{table}

\highlight{For comparison purposes, we also tested the low-rank PSF estimation model described in~\cite{yanny2020miniscope3d, yanny2022deep}. In this model, we use ray tracing to calculate the PSF of the surrounding 8 positions and employ trilinear interpolation to obtain the center PSF. We divide the object space into 20 depths, with 64 grids in each depth plane. The PSFs of these positions are calculated and used for querying. As depicted in Fig.~\ref{fig:psf_pred}, the low-rank PSF model can estimate PSFs similar to the ground truth. Since the PSF is slowly varying in the object space, using enough sampled PSFs for querying can yield promising results.

For a more detailed evaluation, we selected 20 focus distances, 40 depths, and 8 (height) $\times$ 10 (width) positions per depth as the testing dataset to quantify PSF estimation accuracy. Additionally, we compare our proposed network model with the model proposed by Tseng et al.\cite{tseng2021differentiable}. The evaluation metrics used are the $\ell_1$ and $\ell_2$ errors. As shown in Table~\ref{tab:psf_comparison}, our MLP network achieves the most accurate PSF estimation among the three methods. Furthermore, our PSF network can estimate the PSF given discrete spatial points and focus distances without requiring the querying of external data or processing the network output. This advantage makes our network model suitable for aberrated and focused image simulation.}

\section{Aberration-Aware Training Results}

After evaluating the fitting accuracy of the PSF network, we now turn
to evaluate the impact of using this network for aberration-aware
training of the DfF network. To this end, we conduct experiments on both simulated and real-world data. In the first experiment, we train and test the model on simulated focal stacks. In the second experiment, we train the model on simulated focal stacks and test it on real-world focal stacks. Following the AAT scheme, we simulate the training focal stacks with the same lens that is used to simulate/capture the test data.

For comparison, we simulate training focal stacks using the Gaussian PSF calculated by the thin lens model (``non-AAT''). This serves as the baseline, as used in the existing works. To ensure a fair comparison, the thin lens is set to have the same focal length, F-number, and sensor size as the objective real lens, with the only variable being the off-axis optical aberrations. In the experiment, we select two DfF networks: AiFNet~\cite{wang2021bridging} and DFVNet~\cite{yang2022deep}, and train them with the same settings as described in the original papers. For each network, we train two models, one with the AAT scheme and one without it, and evaluate their performance on the testing focal stacks without fine-tuning.

\subsection{Implementation Details}
In the first experiment, we use a 50mm F/2.8 lens\footnote{The lens data comes from \href{http://lensnet.herokuapp.com/}{lensnet.com~\cite{cote2021deep}}, the actual F-number we calculate is F/1.86} (see Fig.~\ref{fig:mlp}(a)), which has significant off-axis aberrations. The image sensor has a physical size of $24{\rm mm} \times 32{\rm mm}$ and a resolution of $480\times 640$, and we train a new PSF network to model this lens. The FlyingThings3D dataset~\cite{mayer2016large, wang2021bridging} is used as the training dataset, which contains 800 pairs of synthetic RGBD images. The Middlebury2014 dataset~\cite{scharstein2014high} is used as the testing dataset, which contains 23 real-world RGBD images. There is a significant domain gap between the synthetic and real-world images, making it suitable for evaluating the generalizability of the DfF models. We simulate focal stacks with the RGBD images for training and testing, using a stack size of 10. The focus distances are chosen linearly from the minimum (20~cm) and maximum (20~m) depth range of each image, with a random perturbation. 

We utilize the AdamW optimizer~\cite{loshchilov2017decoupled} and the CosineAnnealing learning rate scheduler~\cite{loshchilov2016sgdr} with their default parameters. The training batch size is set to 16, and the initial learning rate is set to $1e^{-4}$. Each DfF model is trained for 400 epochs on a single A100 80G GPU which is enough for convergence. After training, we evaluate each model on the testing dataset without fine-tuning.

\begin{table*}[ht]
    \caption{\textbf{Quantitative depth estimation results on simulated focal stacks.} Two DfF networks, AiFNet~\cite{wang2021bridging} and DFVNet~\cite{yang2022deep}, are chosen for depth estimation. The testing focal stacks are rendered by real lens PSF, while we use Gaussian PSF (``Baseline'') and real lens PSF (``Ours'') to render training focal stacks. Models trained with our AAT scheme deliver better depth estimation accuracy.}
    \centering
    \renewcommand{\arraystretch}{1.5}
    \resizebox{\textwidth}{!}{
    \begin{tabular}{l||c|c|c|c|c|c|c|c}
        \hline
        Method & MAE $\downarrow$ & MSE $\downarrow$ & RMSE $\downarrow$ & Abs. rel. $\downarrow$ & Sqr. rel. $\downarrow$ & $\delta =1.25\uparrow$ & $\delta=1.25^{2} \uparrow$ & $\delta=1.25^{3} \uparrow$\\
        \hline
        Baseline (AiFNet) & 0.4706 & 0.4805 & 0.6229 & 0.2759 & 0.4906 & 0.8098 & 0.9587 & 0.9713 \\
        Ours (AiFNet) & \textbf{0.2095} & \textbf{0.1536} & \textbf{0.3475} & \textbf{0.1613} & \textbf{0.2982} & \textbf{0.9683} & \textbf{0.9852} & \textbf{0.9895} \\
        \hline
        Baseline (DFVNet) & 0.5900 & 0.8452 & 0.8025 & 0.2885 & 0.5065 & 0.7390 & 0.8903 & 0.9277 \\
        Ours (DFVNet) & \textbf{0.1977} & \textbf{0.1582} & \textbf{0.3446} & \textbf{0.1563} & \textbf{0.2992} & \textbf{0.9669} & \textbf{0.9830} & \textbf{0.9876} \\
        \hline
    \end{tabular}
    }
    \label{tab:comparison}
\end{table*}

\begin{figure*}[!ht]
    \centering
    \setlength{\tabcolsep}{1pt}
    \resizebox{\textwidth}{!}{
    \begin{tabular}{ccccc}
         GT & Ours (AiFNet) & Baseline (AiFNet) & Ours (DFVNet) & Baseline (DFVNet)
         \\
         \includegraphics[width=3cm]{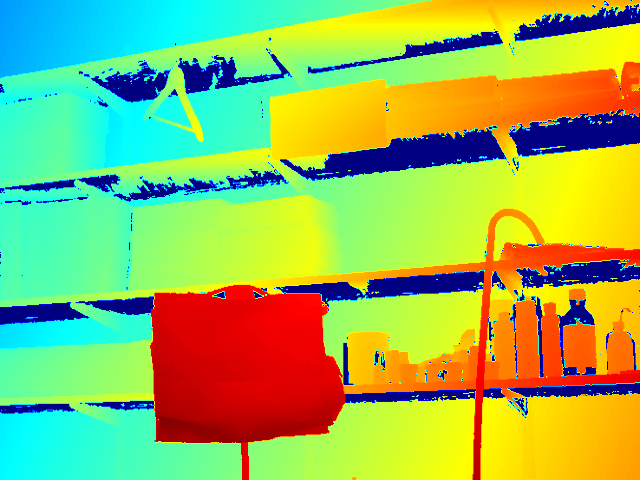}
         &
         \includegraphics[width=3cm]{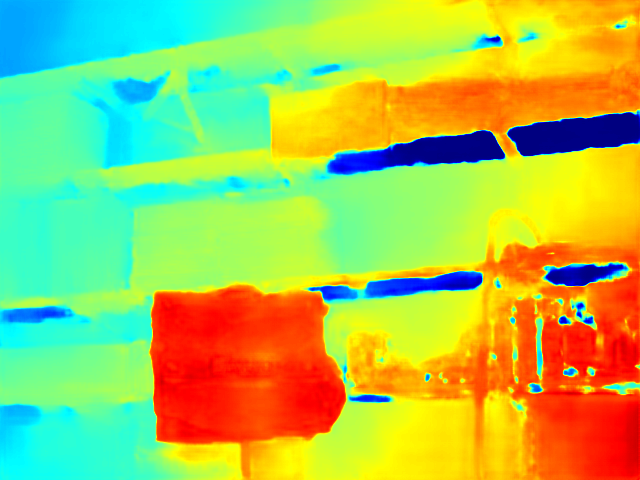}
         & 
         \includegraphics[width=3cm]{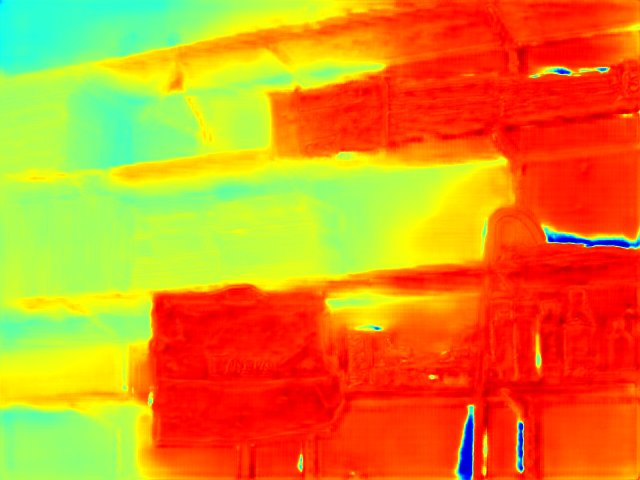}
         &
         \includegraphics[width=3cm]{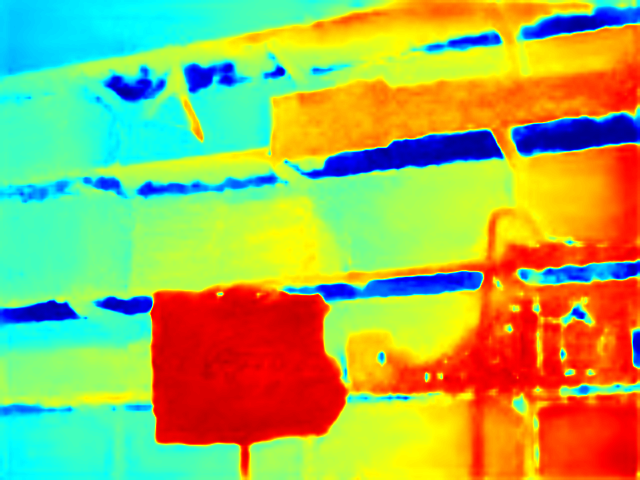}
         &
         \includegraphics[width=3cm]{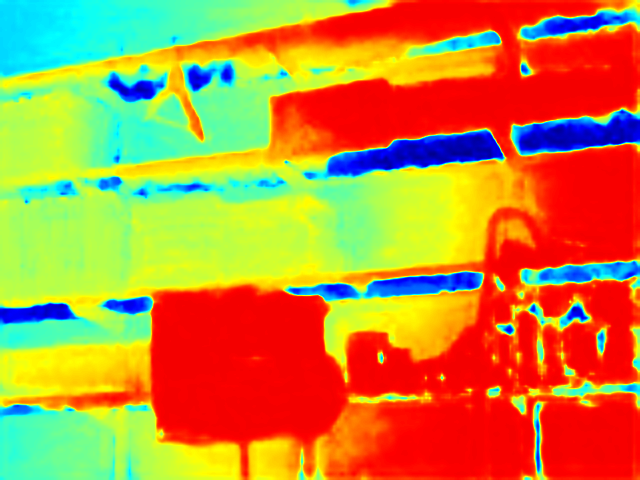}
         \\
         \includegraphics[width=3cm]{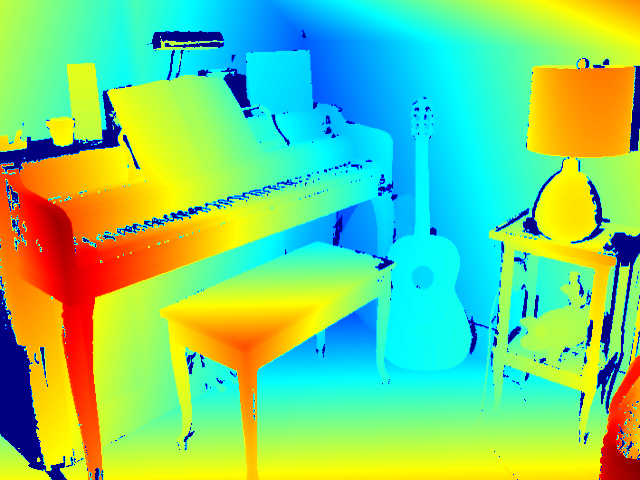}
         &
         \includegraphics[width=3cm]{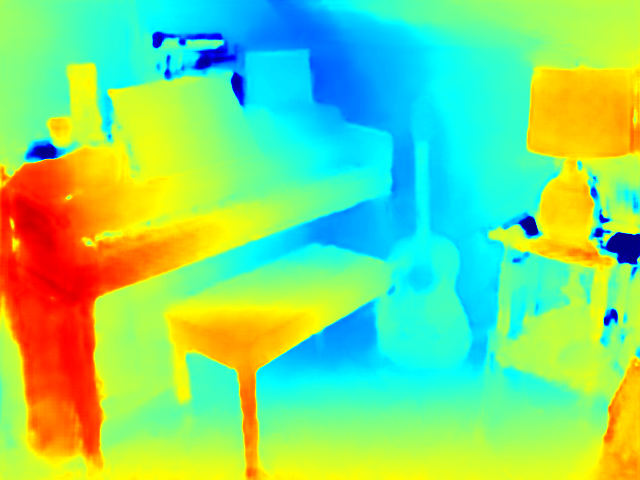}
         & 
         \includegraphics[width=3cm]{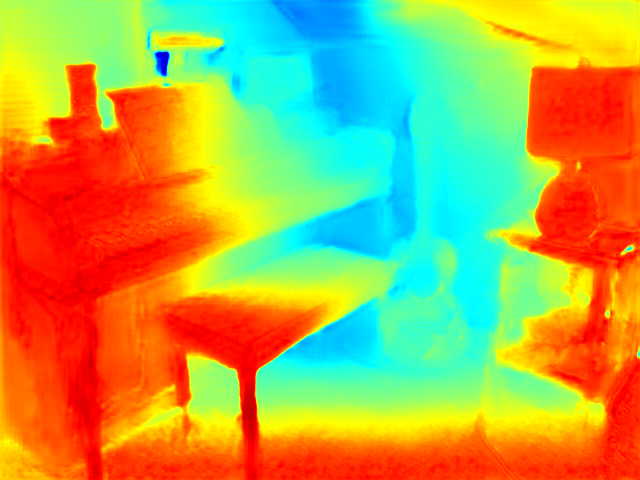}
         &
         \includegraphics[width=3cm]{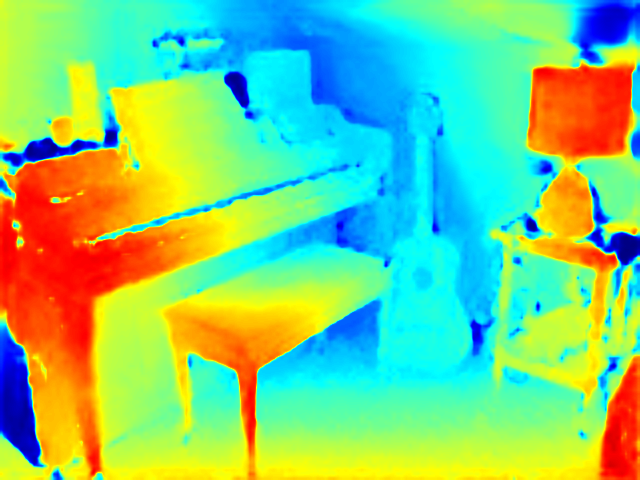}
         &
         \includegraphics[width=3cm]{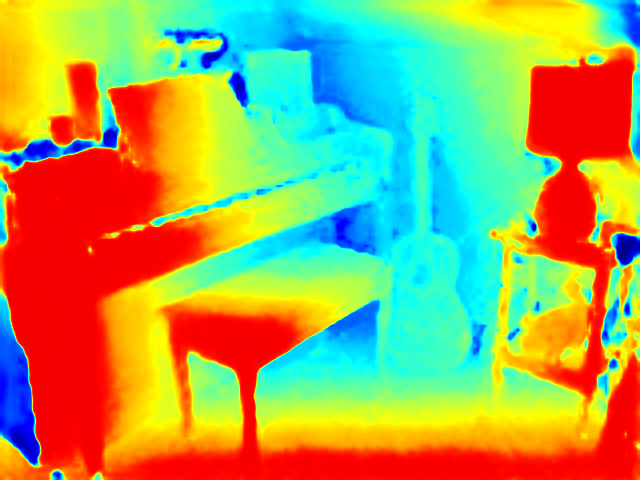}
         \\
         \includegraphics[width=3cm]{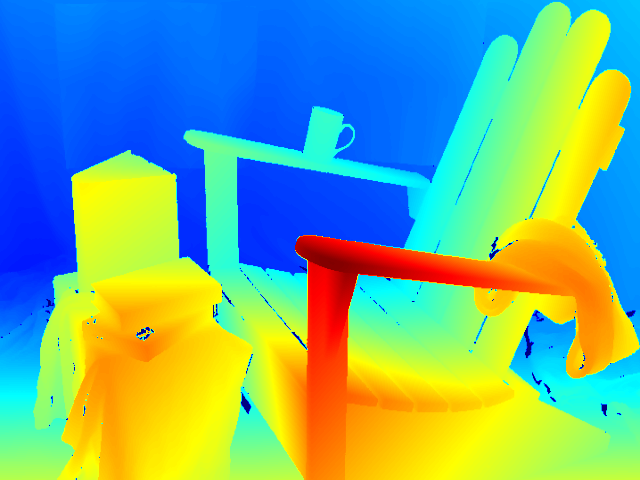}
         &
         \includegraphics[width=3cm]{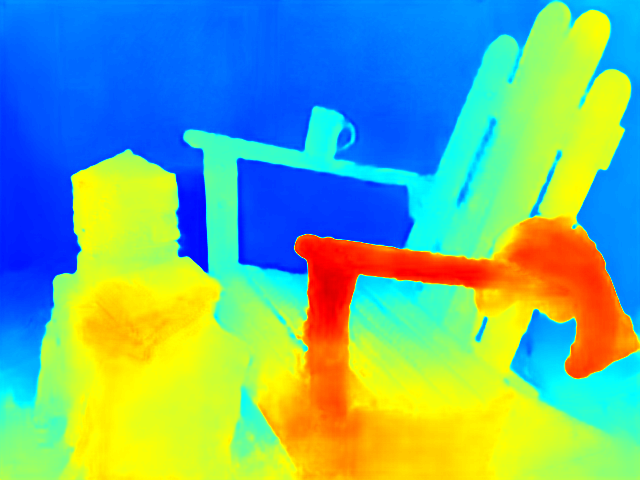}
         & 
         \includegraphics[width=3cm]{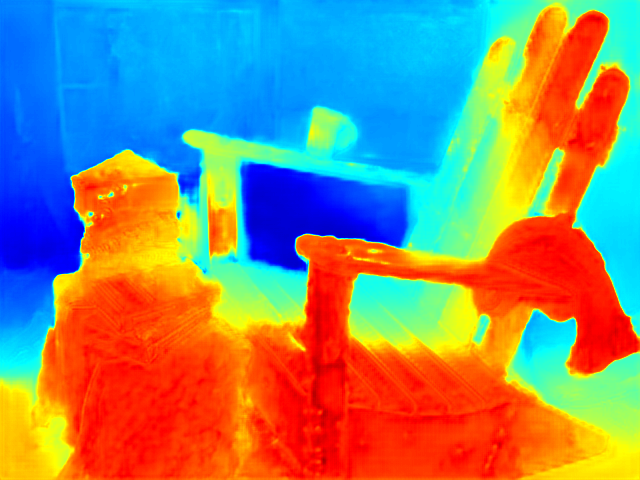}
         &
         \includegraphics[width=3cm]{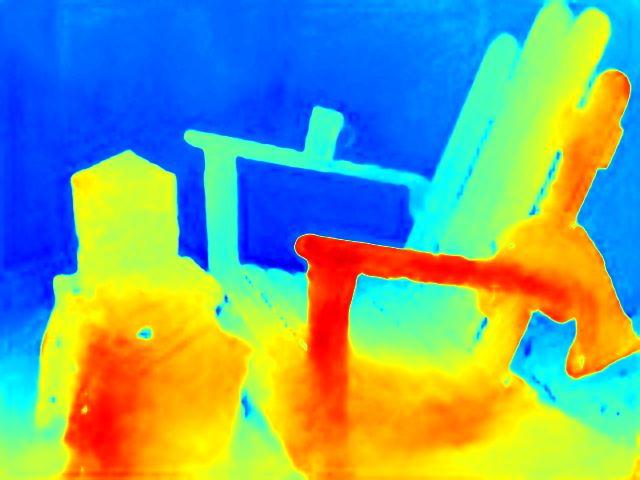}
         &
         \includegraphics[width=3cm]{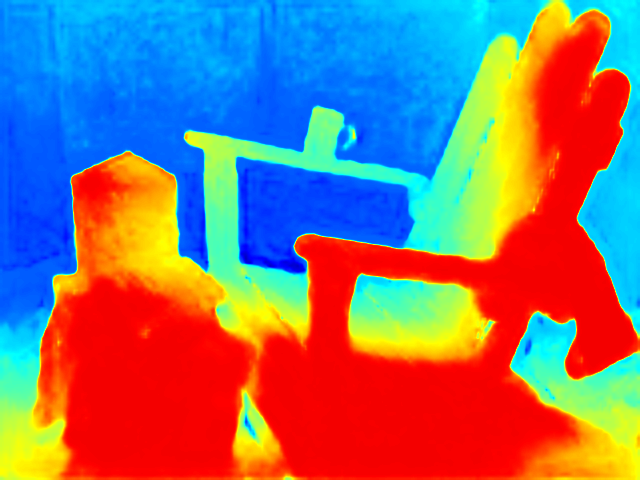}
    \end{tabular}
    }
    \caption{\textbf{Qualitative results on simulated focal stacks.} With the AAT scheme, both network models predict more accurate and finer depth maps, while the non-AAT models fail to distinguish between adjacent objects and also mispredict the depth of some edge objects.}
    \label{fig:comparison}
\end{figure*}

In the second experiment, we use the Canon EOS R camera and the RF 50mm F/1.8 lens, discussed in Section~\ref{sec:psf-rep}. The DfF training procedure remains the same as the previous experiment. We use the Matterport3D dataset preprocessed by Zhang et al.~\cite{zhang2018deep}, which comprises 117,516 pairs of indoor RGBD images as additional training data and train each model for an additional 20 epochs, followed by an evaluation of real-world captured focal stacks. The real-world captured focal stacks consist of 24 outdoor scenes and 1 indoor scene.

\begin{figure*}[ht]
    \centering
    \setlength{\tabcolsep}{1pt}
    \resizebox{\linewidth}{!}{
    \begin{tabular}{cccccc}
         % GT & Ours (AiFNet) & Baseline (AiFNet)
         % \\
         \rotatebox{90}{\footnotesize \quad \quad \quad GT}
         &
         \begin{tikzpicture}[spy using outlines={rectangle, magnification=3, size=1.0cm}]
            \node[anchor=south west, inner sep=0] (image) at (0,0) {\includegraphics[width=3cm]{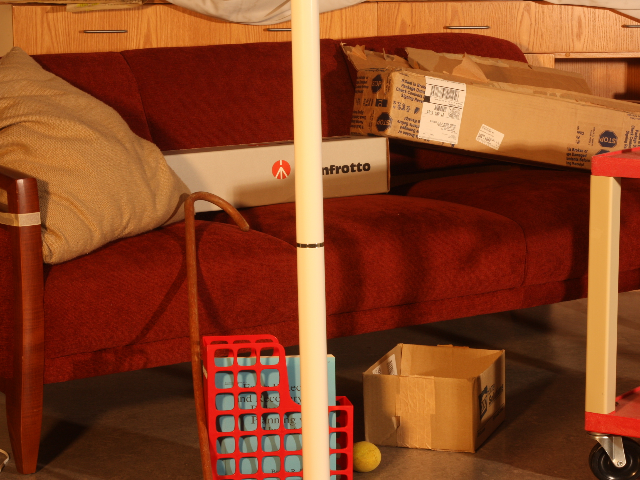}};
            \spy[white, solid, thick] on (1.6, 0.4) in node [below] at (2.5, 2.25);
         \end{tikzpicture}
         &
         \begin{tikzpicture}[spy using outlines={rectangle, magnification=3, size=1.0cm}]
            \node[anchor=south west, inner sep=0] (image) at (0,0) {\includegraphics[width=3cm]{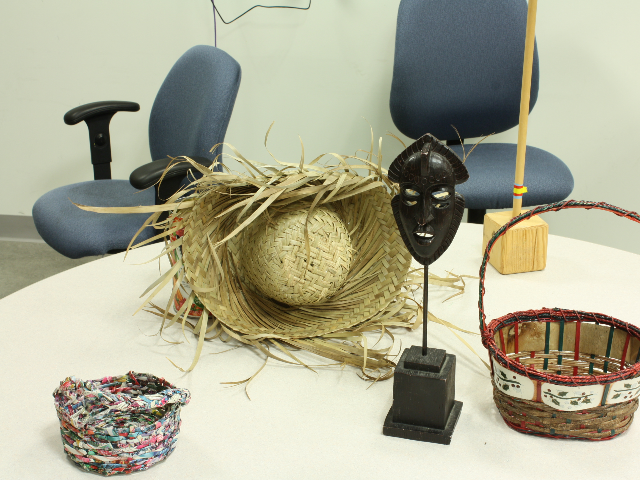}};
            \spy[white, solid, thick] on (0.65, 1.4) in node [below] at (2.5, 2.25);
         \end{tikzpicture}
         & 
         \begin{tikzpicture}[spy using outlines={rectangle, magnification=3, size=1.0cm}]
            \node[anchor=south west, inner sep=0] (image) at (0,0) {\includegraphics[width=3cm]{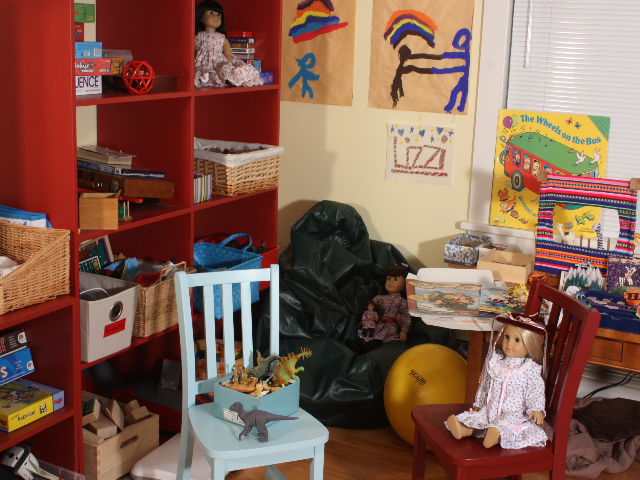}};
            \spy[white, solid, thick] on (1.25, 0.8) in node [below] at (2.5, 2.25);
         \end{tikzpicture}
         &
         \begin{tikzpicture}[spy using outlines={rectangle, magnification=3, size=1.0cm}]
            \node[anchor=south west, inner sep=0] (image) at (0,0) {\includegraphics[width=3cm]{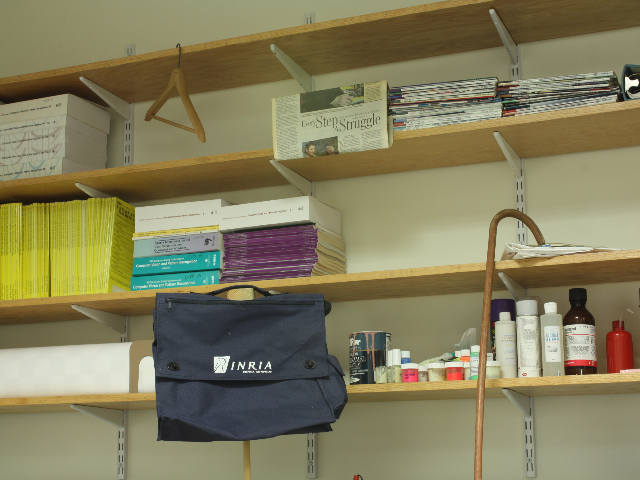}};
            % \spy[white, solid, thick] on (1.82, 1.5) in node [below] at (2.5, 2.25);
            \spy[white, solid, thick] on (2.7, 0.5) in node [below] at (2.5, 2.25);
         \end{tikzpicture}
         & 
         \begin{tikzpicture}[spy using outlines={rectangle, magnification=3, size=1.0cm}]
            \node[anchor=south west, inner sep=0] (image) at (0,0) {\includegraphics[width=3cm]{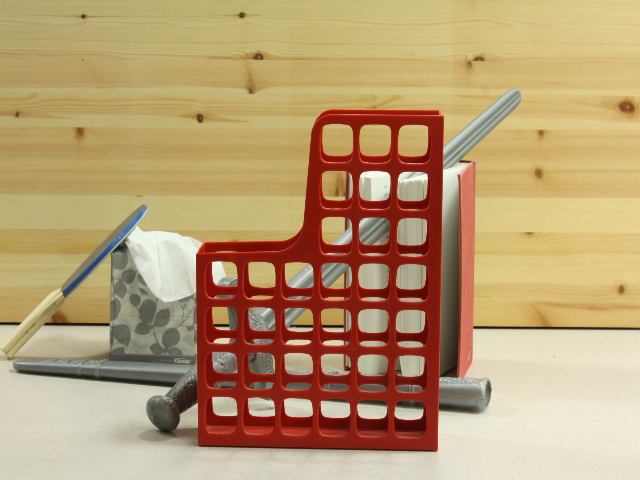}};
            \spy[white, solid, thick] on (1.35, 0.55) in node [below] at (2.5, 2.25);
         \end{tikzpicture}
         \\
         \rotatebox{90}{\footnotesize \quad Ours (AiFNet)}
         &
         \begin{tikzpicture}[spy using outlines={rectangle, magnification=3, size=1.0cm}]
            \node[anchor=south west, inner sep=0] (image) at (0,0) {\includegraphics[width=3cm]{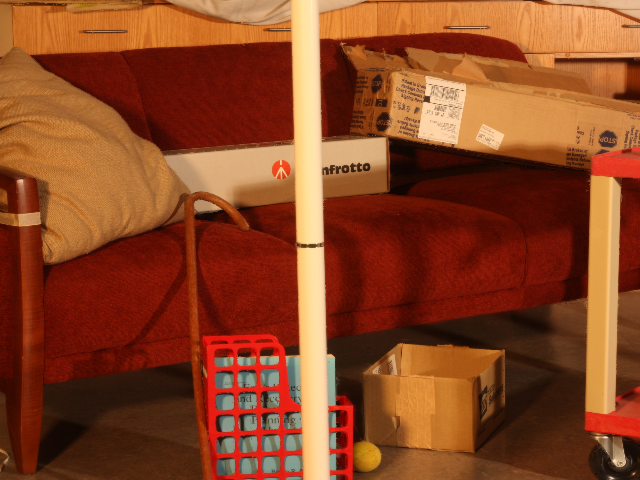}};
            \spy[white, solid, thick] on (1.6, 0.4) in node [below] at (2.5, 2.25);
         \end{tikzpicture}
         &
         \begin{tikzpicture}[spy using outlines={rectangle, magnification=3, size=1.0cm}]
            \node[anchor=south west, inner sep=0] (image) at (0,0) {\includegraphics[width=3cm]{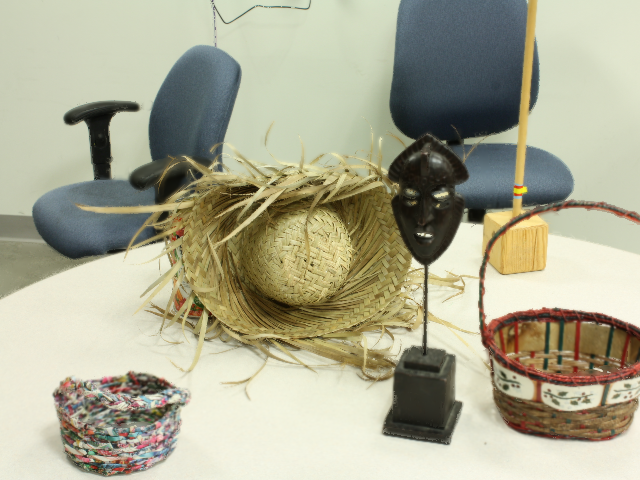}};
            \spy[white, solid, thick] on (0.65, 1.4) in node [below] at (2.5, 2.25);
         \end{tikzpicture}
         & 
         \begin{tikzpicture}[spy using outlines={rectangle, magnification=3, size=1.0cm}]
            \node[anchor=south west, inner sep=0] (image) at (0,0) {\includegraphics[width=3cm]{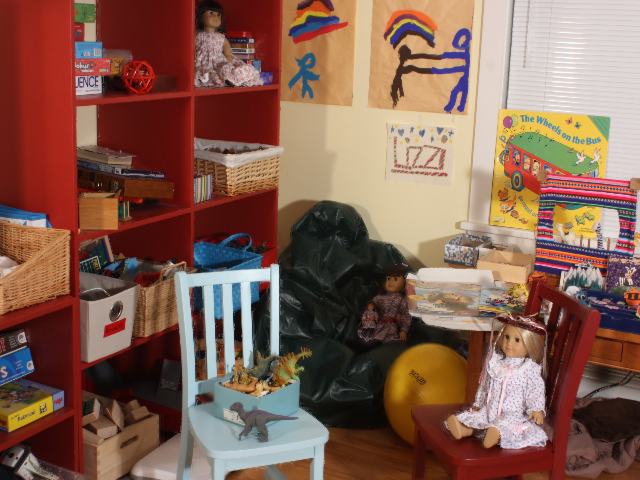}};
            \spy[white, solid, thick] on (1.25, 0.8) in node [below] at (2.5, 2.25);
         \end{tikzpicture}
         & 
         \begin{tikzpicture}[spy using outlines={rectangle, magnification=3, size=1.0cm}]
            \node[anchor=south west, inner sep=0] (image) at (0,0) {\includegraphics[width=3cm]{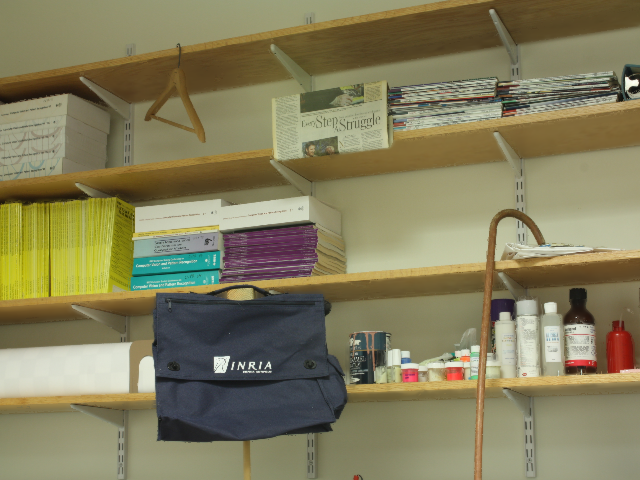}};
            % \spy[white, solid, thick] on (1.82, 1.5) in node [below] at (2.5, 2.25);
            \spy[white, solid, thick] on (2.7, 0.5) in node [below] at (2.5, 2.25);
         \end{tikzpicture}
         & 
         \begin{tikzpicture}[spy using outlines={rectangle, magnification=3, size=1.0cm}]
            \node[anchor=south west, inner sep=0] (image) at (0,0) {\includegraphics[width=3cm]{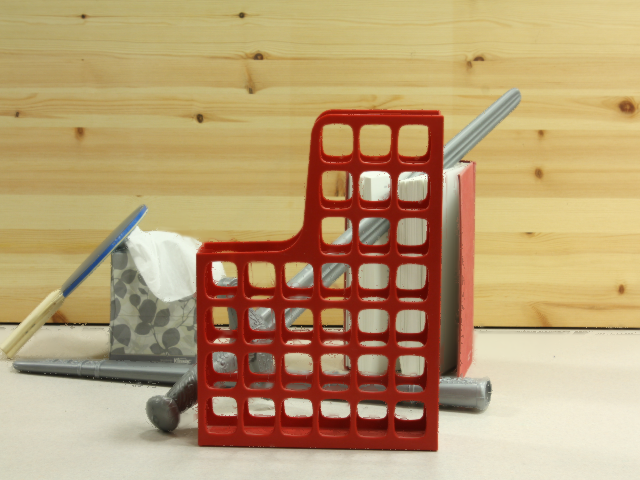}};
            \spy[white, solid, thick] on (1.35, 0.55) in node [below] at (2.5, 2.25);
         \end{tikzpicture}
         \\
         \rotatebox{90}{\footnotesize Baseline (AiFNet)}
         &
         \begin{tikzpicture}[spy using outlines={rectangle, magnification=3, size=1.0cm}]
            \node[anchor=south west, inner sep=0] (image) at (0,0) {\includegraphics[width=3cm]{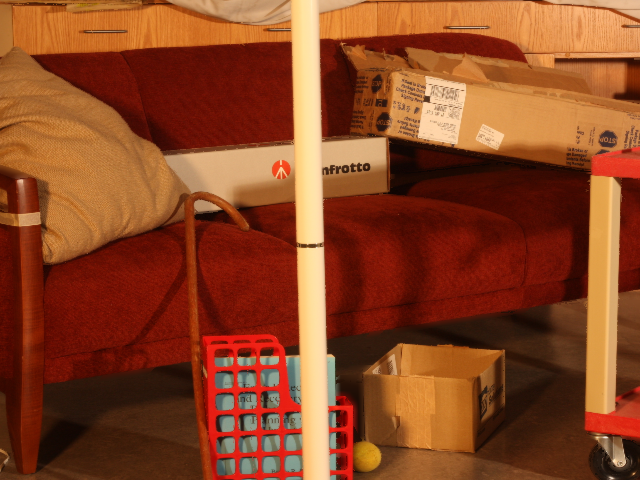}};
            \spy[white, solid, thick] on (1.6, 0.4) in node [below] at (2.5, 2.25);
         \end{tikzpicture}
         &
         \begin{tikzpicture}[spy using outlines={rectangle, magnification=3, size=1.0cm}]
            \node[anchor=south west, inner sep=0] (image) at (0,0) {\includegraphics[width=3cm]{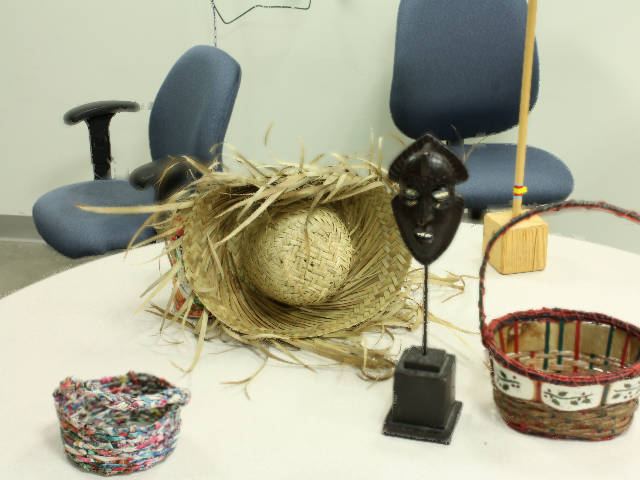}};
            \spy[white, solid, thick] on (0.65, 1.4) in node [below] at (2.5, 2.25);
         \end{tikzpicture}
         & 
         \begin{tikzpicture}[spy using outlines={rectangle, magnification=3, size=1.0cm}]
            \node[anchor=south west, inner sep=0] (image) at (0,0) {\includegraphics[width=3cm]{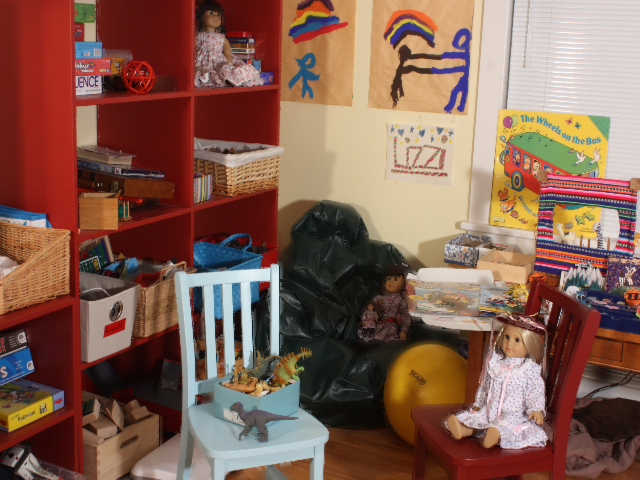}};
            \spy[white, solid, thick] on (1.25, 0.8) in node [below] at (2.5, 2.25);
         \end{tikzpicture}
         & 
         \begin{tikzpicture}[spy using outlines={rectangle, magnification=3, size=1.0cm}]
            \node[anchor=south west, inner sep=0] (image) at (0,0) {\includegraphics[width=3cm]{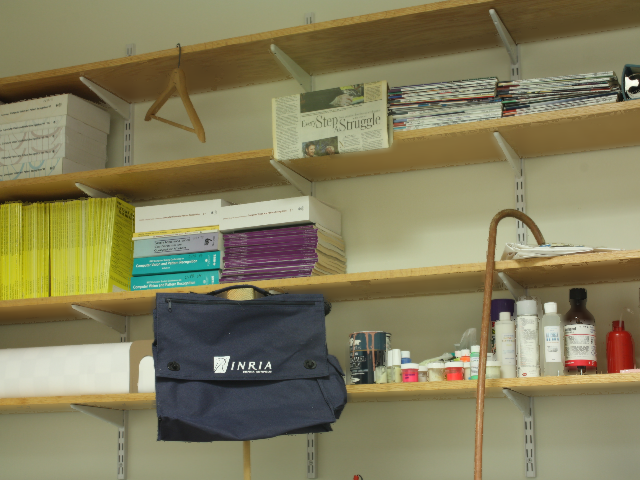}};
            % \spy[white, solid, thick] on (1.82, 1.5) in node [below] at (2.5, 2.25);
            \spy[white, solid, thick] on (2.7, 0.5) in node [below] at (2.5, 2.25);
         \end{tikzpicture}
         & 
         \begin{tikzpicture}[spy using outlines={rectangle, magnification=3, size=1.0cm}]
            \node[anchor=south west, inner sep=0] (image) at (0,0) {\includegraphics[width=3cm]{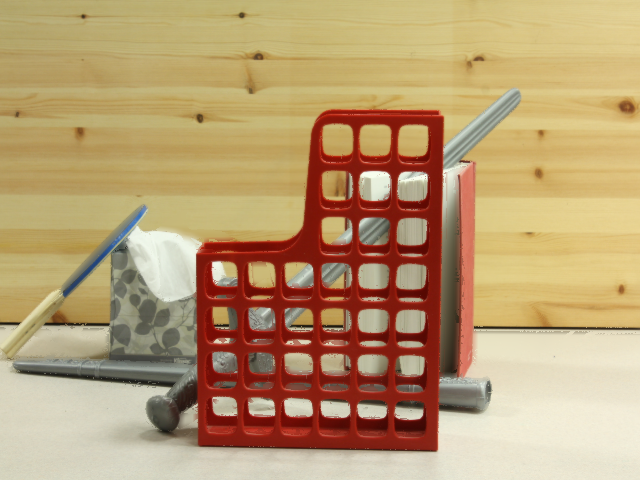}};
            \spy[white, solid, thick] on (1.35, 0.55) in node [below] at (2.5, 2.25);
         \end{tikzpicture}
    \end{tabular}
    }
    \caption{\textbf{All-in-focus image synthesis results on simulated focal stacks.} The all-in-focus images synthesized without the AAT scheme exhibit significant estimation errors at the edge of the objects, and the continuous image regions show inconsistent sharpness and blurriness. In contrast, the AAT scheme results in clear and continuous estimated edges.}
    \label{fig:comparison_aif}
\end{figure*}

\subsection{Evaluation on Simulated Focal Stacks}
\subsubsection{Depth estimation}
In Fig.~\ref{fig:comparison}, we present the qualitative results of the estimated depth maps. Despite the significant domain gap between the original synthetic training data and the real-world test data, both DfF models can estimate promising depth maps with the AAT scheme. This is because we simulate both training and testing focal stacks with the same lens model, and the optical aberrations are independent of semantic information in the images. We observe that small depth differences between adjacent objects are preserved in the estimated depth maps. Furthermore, the absolute depth values of the objects are precisely estimated without significant errors. 

However, the depth maps estimated by the non-AAT models fail to distinguish between adjacent objects, and also there are errors at the corner of the depth maps. This inaccuracy is caused by the off-axis aberrations, which is the only variable in the experiment. As illustrated in the previous section, the off-axis aberrations, such as field curvature, affects the decision of the best in-focus frame in the focus stack and degrade the performance of pre-trained models. These off-axis aberrations have a dominant effect at smaller out-of-focus depths, leading to the non-AAT model's inability to discriminate between adjacent objects.

In Table.~\ref{tab:comparison}, we evaluate quantitative results with the following metrics: mean-absolute error (MAE), mean-squared error (MSE), root-mean-squared error (RMSE), relative-absolute error (Abs.rel.), relative-squared error (Sqr.rel.), accuracy with $\delta = 1.25$. For MAE, MSE, RMSE, Abs.rel. and Sqr.rel., lower values indicate better results, while for the three accuracy metrics, higher values represent better results. As shown in the table, both AiFNet and DFVNet models greatly improve with the AAT compared to the non-AAT results.
% \WH{am I missing something? I don't see logRMS anywhere, and anyway it
% would be redundant information with the RMS in the sense that there is
% a one-to-one mapping between the two... In general I think you might
% want to strip redundant metrics...}

\subsubsection{All-in-Focus Image Synthesis}

\highlight{We also evaluated the results of all-in-focus image synthesis, which utilized the same predicted probability maps but instead interpolated RGB values of focused images for the final output. The synthesized all-in-focus images are presented in Fig.~\ref{fig:comparison_aif}, with zoomed image patches in the top-right corner. As observed, the edges of the objects in the synthesized images without the AAT scheme exhibit a fuzzy effect. Furthermore, there are noticeable estimation errors that the continuous image regions exhibit inconsistent sharpness and blurriness. PSNR and SSIM scores of estimated all-in-focus images are presented in Table~\ref{tab:aif_comparison}, the AAT scheme leads to higher all-in-focus image synthesis quality compared to the baseline.}

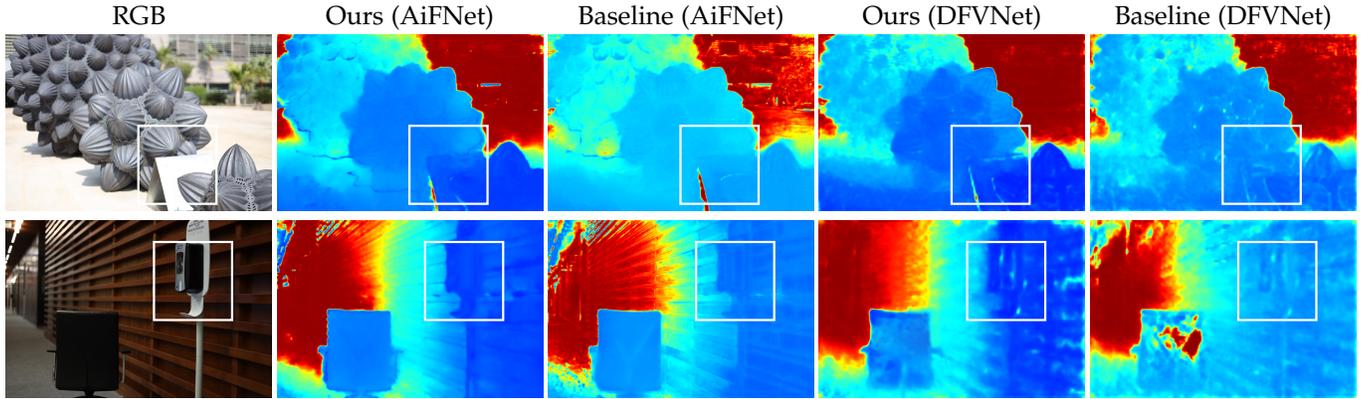
\begin{figure*}[htbp]
    \centering

    \setlength{\tabcolsep}{1pt}
    \resizebox{\textwidth}{!}{
    \begin{tabular}{ccccc}
            RGB & Ours (AiFNet) & Baseline (AiFNet) & Ours (DFVNet) & Baseline (DFVNet)
         \\
            \begin{tikzpicture}
                \node[anchor=south west,inner sep=0] at (0,0) {\includegraphics[width=3.4cm]{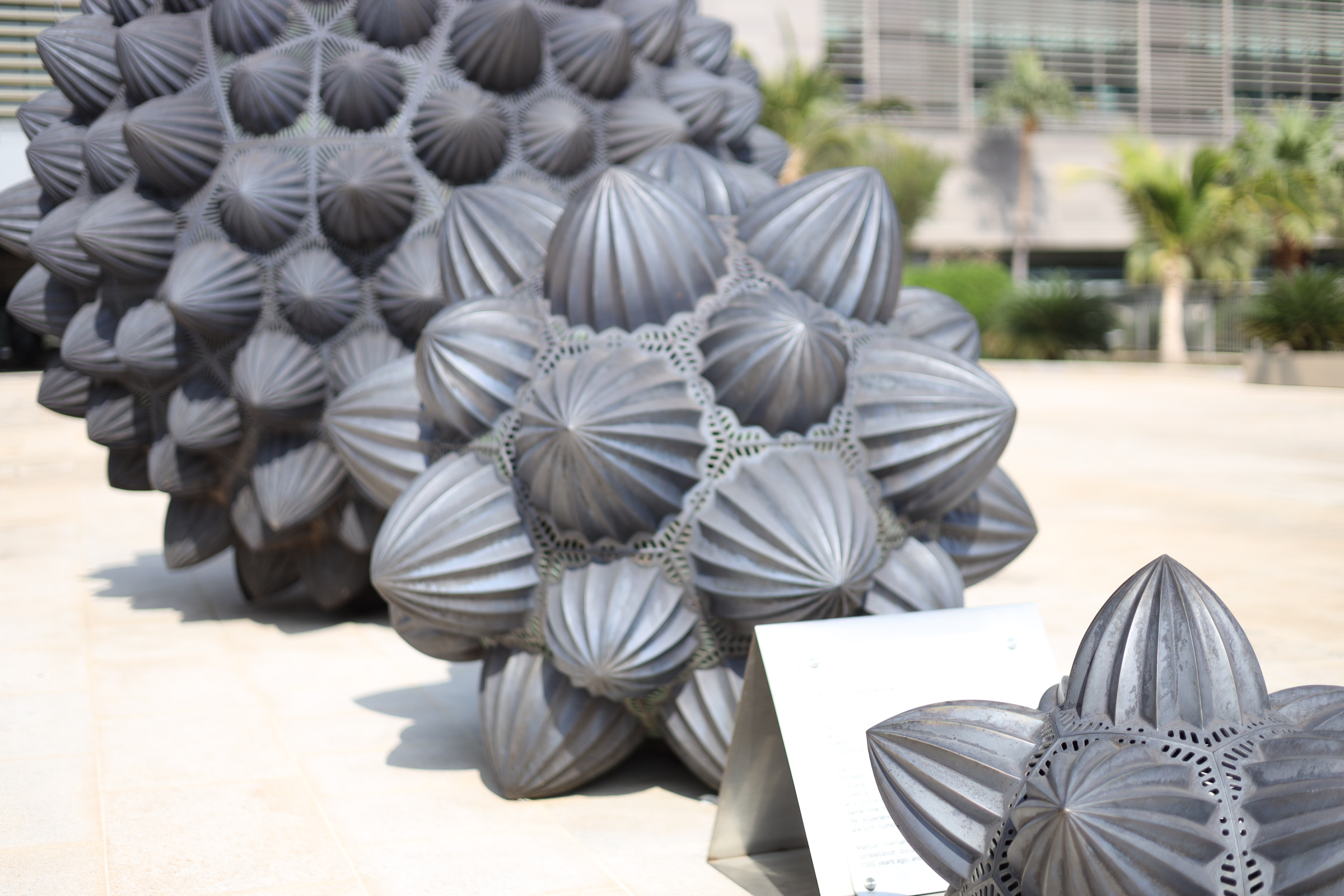}};
                \draw[white,thick] (1.7, 1.1) rectangle (2.7, 0.1);
            \end{tikzpicture}
             
             &
             \begin{tikzpicture}
                \node[anchor=south west,inner sep=0] at (0,0) {\includegraphics[width=3.4cm]{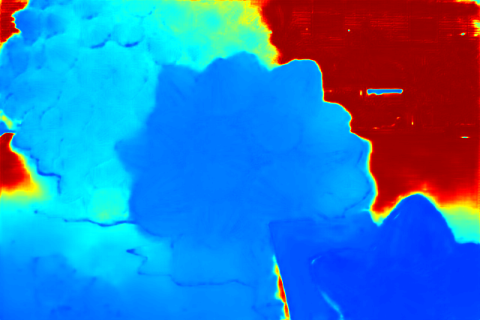}};
                \draw[white,thick] (1.7, 1.1) rectangle (2.7, 0.1);
            \end{tikzpicture}
             & 
             \begin{tikzpicture}
                \node[anchor=south west,inner sep=0] at (0,0) {\includegraphics[width=3.4cm]{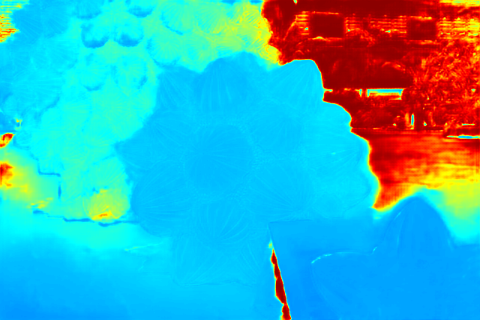}};
                \draw[white,thick] (1.7, 1.1) rectangle (2.7, 0.1);
            \end{tikzpicture}
             &
             \begin{tikzpicture}
                \node[anchor=south west,inner sep=0] at (0,0) {\includegraphics[width=3.4cm]{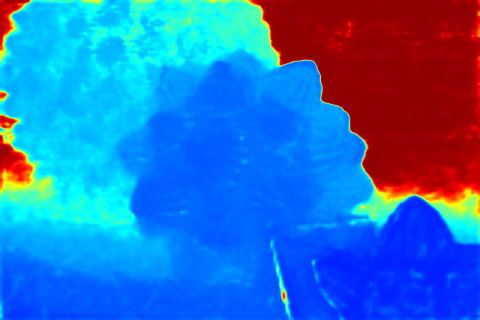}};
                \draw[white,thick] (1.7, 1.1) rectangle (2.7, 0.1);
            \end{tikzpicture}
             & 
             \begin{tikzpicture}
                \node[anchor=south west,inner sep=0] at (0,0) {\includegraphics[width=3.4cm]{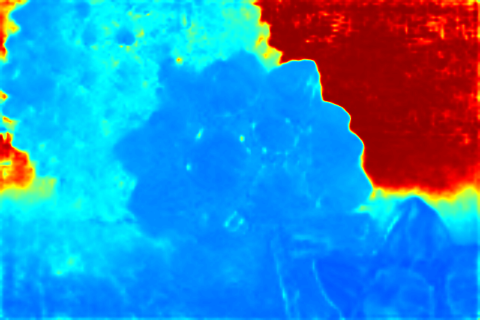}};
                \draw[white,thick] (1.7, 1.1) rectangle (2.7, 0.1);
            \end{tikzpicture}
         \\
            \begin{tikzpicture}
                \node[anchor=south west,inner sep=0] at (0,0) {\includegraphics[width=3.4cm]{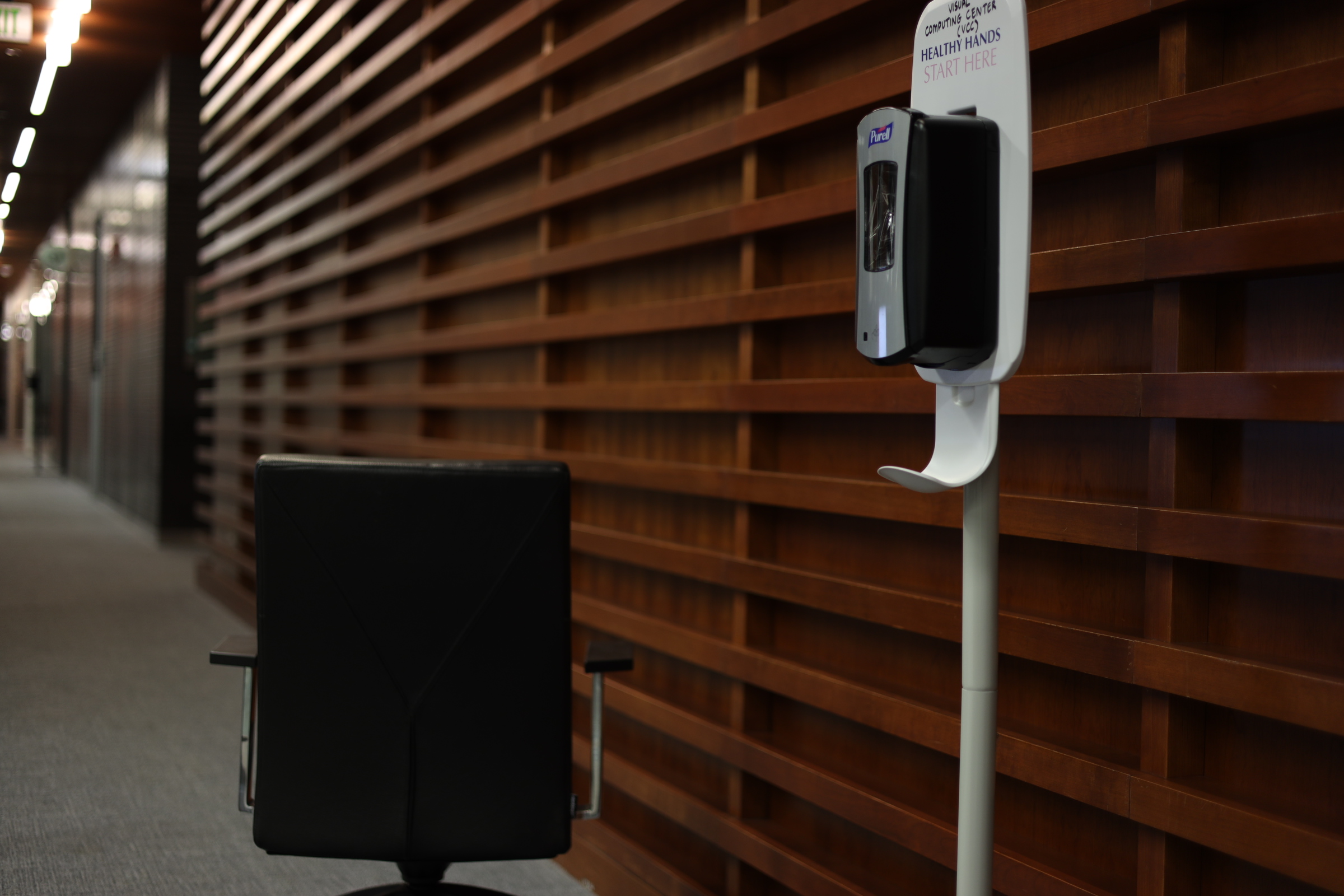}};
                \draw[white,thick] (1.9,2.0) rectangle (2.9, 1.0);
            \end{tikzpicture}
            &
            \begin{tikzpicture}
                \node[anchor=south west,inner sep=0] at (0,0) {\includegraphics[width=3.4cm]{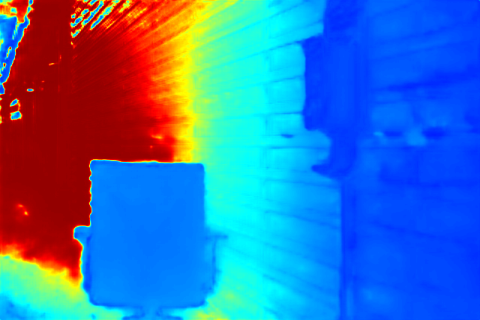}};
                \draw[white,thick] (1.9,2.0) rectangle (2.9, 1.0);
            \end{tikzpicture}
            & 
            \begin{tikzpicture}
                \node[anchor=south west,inner sep=0] at (0,0) {\includegraphics[width=3.4cm]{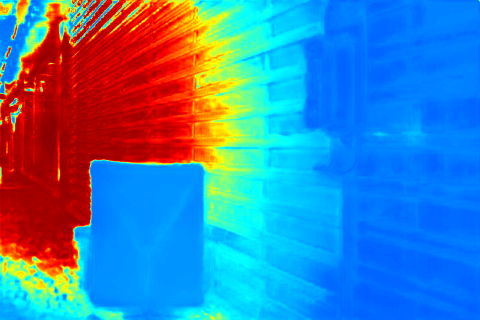}};
                \draw[white,thick] (1.9,2.0) rectangle (2.9, 1.0);
            \end{tikzpicture}
            &
            \begin{tikzpicture}
                \node[anchor=south west,inner sep=0] at (0,0) {\includegraphics[width=3.4cm]{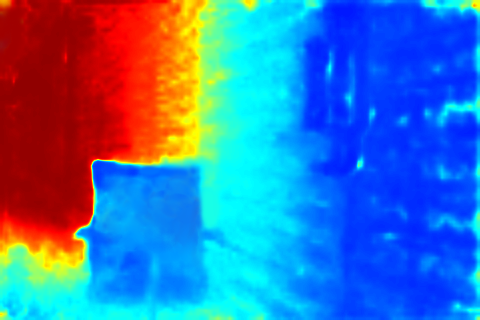}};
                \draw[white,thick] (1.9,2.0) rectangle (2.9, 1.0);
            \end{tikzpicture}
            & 
            \begin{tikzpicture}
                \node[anchor=south west,inner sep=0] at (0,0) {\includegraphics[width=3.4cm]{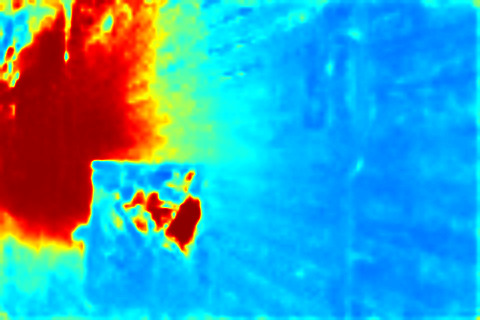}};
                \draw[white,thick] (1.9,2.0) rectangle (2.9, 1.0);
            \end{tikzpicture}
    \end{tabular}
    }
    \caption{\textbf{Qualitative results on real-world outdoor focal stacks.} We capture real-world focal stacks using the same camera lens that is used for training and test pretrained models on them. The depth map predicted by AAT models is more hierarchical and can better distinguish depth differences between neighboring objects, as labeled by boxes.}
    \label{fig:realworld_outdoor}
\end{figure*}

\begin{table}[ht]
    \centering
    \caption{Quantitative evaluation of AAT scheme on all-in-focus image synthesis.}
    \renewcommand{\arraystretch}{1.2}
    \resizebox{0.55\linewidth}{!}{ 
    \begin{tabular}{l||c|c}
        \hline
        AiFNet & PSNR & SSIM \\
        \hline
        Baseline & 33.55 & 0.970 \\
        \hline
        Ours & \textbf{34.65} & \textbf{0.976} \\
        \hline
    \end{tabular}
    }
    \label{tab:aif_comparison}
\end{table}

\subsection{Evaluation on Real-World Focal Stacks}

We capture real-world focal stacks using the Canon RF 50~mm F/1.8 lens and test pre-trained DfF models. We select 24 outdoor scenes and one indoor scene for evaluation. 

\subsubsection{Outdoor Scenes Evaluation}
The qualitative results of outdoor scenes are presented in Fig.~\ref{fig:realworld_outdoor}, where we obtain the focus distance from the photo's EXIF data. We observe that all models estimate good depth maps, but the AAT-trained models provide finer details and smoother results. Moreover, the AAT-trained models give a more hierarchical depth map and can distinguish depth differences between neighboring objects better. In contrast, the non-AAT models estimate less accurate depth maps, and the depth maps tend to confuse objects with little difference in depth, as indicated by the objects marked by boxes in the images. The two models exhibit significant differences at the corner of the estimated depth maps because off-axis lens aberrations are more pronounced at the image edge than at the center.

\begin{table}[ht]
    \centering
    \caption{Quantitative results on real-world indoor focal stacks.}
    % \footnotesize
    \renewcommand{\arraystretch}{1.6}
    \small
    \begin{tabular}{l||c|c|c|c}
        \hline
        AiFNet & MAE $\downarrow$ & MSE $\downarrow$ & Abs.Rel. $\downarrow$ & $\delta=1.25 \uparrow$
        \\
        \hline
        Baseline & 0.1775 & 0.0836 & 0.1736 & 0.7978 \\
        \hline
        Ours & \textbf{0.1526} & \textbf{0.0770} & \textbf{0.1486} & \textbf{0.9029} \\
        \hline
    \end{tabular}
    \label{tab:realworld_indoor}
\end{table}

\subsubsection{Indoor Scene Evaluation}
\begin{figure}[ht]
    \centering
    \includegraphics[width=\linewidth]{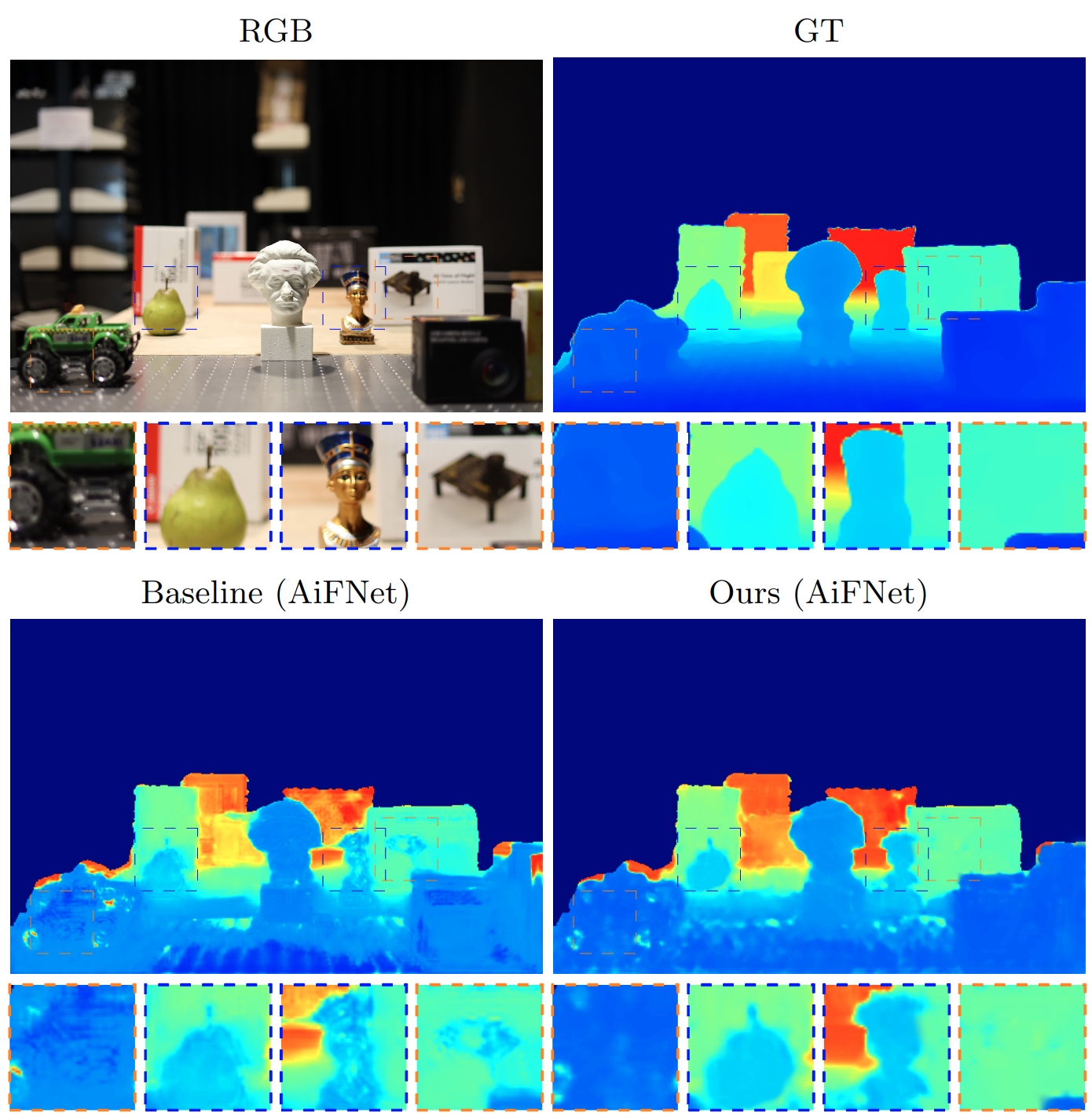}
    \caption{\textbf{Qualitative results on real-world indoor focal stacks.} We set up an indoor scene with objects placed at different positions. The non-AAT model incorrectly treats texture information in RGB images as depth information and also loses edge details in the estimated depth map.}
    \label{fig:realworld_indoor}
\end{figure}

In Fig.~\ref{fig:realworld_indoor}, we present an indoor scene with objects at different positions and depths. We focus the camera on various objects to form the focal stack and measure more accurate focus distances with a ruler. Furthermore, we use the Lidar sensor from the iPhone14 Pro to scan the 3D scene, load the meshes into Blender software, and calibrate the camera's position and view. After calibration, we render the depth map as the ground truth and calculate the score metrics. In Table.~\ref{tab:realworld_indoor}, the quantitative scores indicate that the AAT pre-trained model performs better than the baseline (``non-AAT''). We also observe from the zoomed image patches in Fig.~\ref{fig:realworld_indoor} that the non-AAT model incorrectly treats texture information in RGB images as depth information and loses edge information. For more results on the evaluation, please refer to the Supplementary.

% =============================================================
% =============================================================
\subsection{Efficiency Evaluation of AAT Scheme}

\begin{figure}[ht]
	\small
	\centering
	\setlength{\tabcolsep}{1pt}
        \resizebox{\linewidth}{!}{
	\begin{tabular}{cc}
            {Baseline (AiFNet)}
            &
            {Ours (AiFNet)}
            \\
            \includegraphics[width=0.5\linewidth]{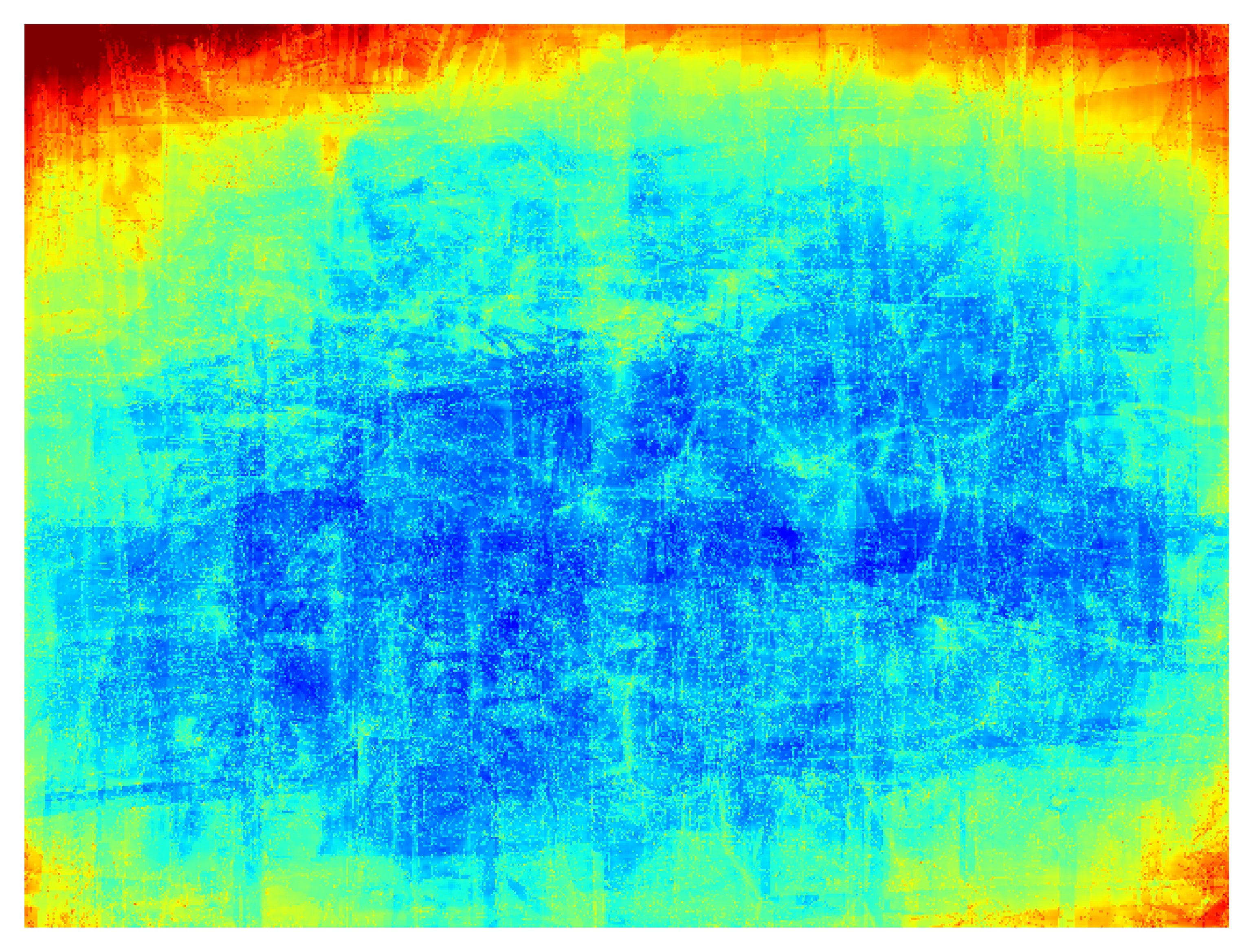}
            &
            \includegraphics[width=0.5\linewidth]{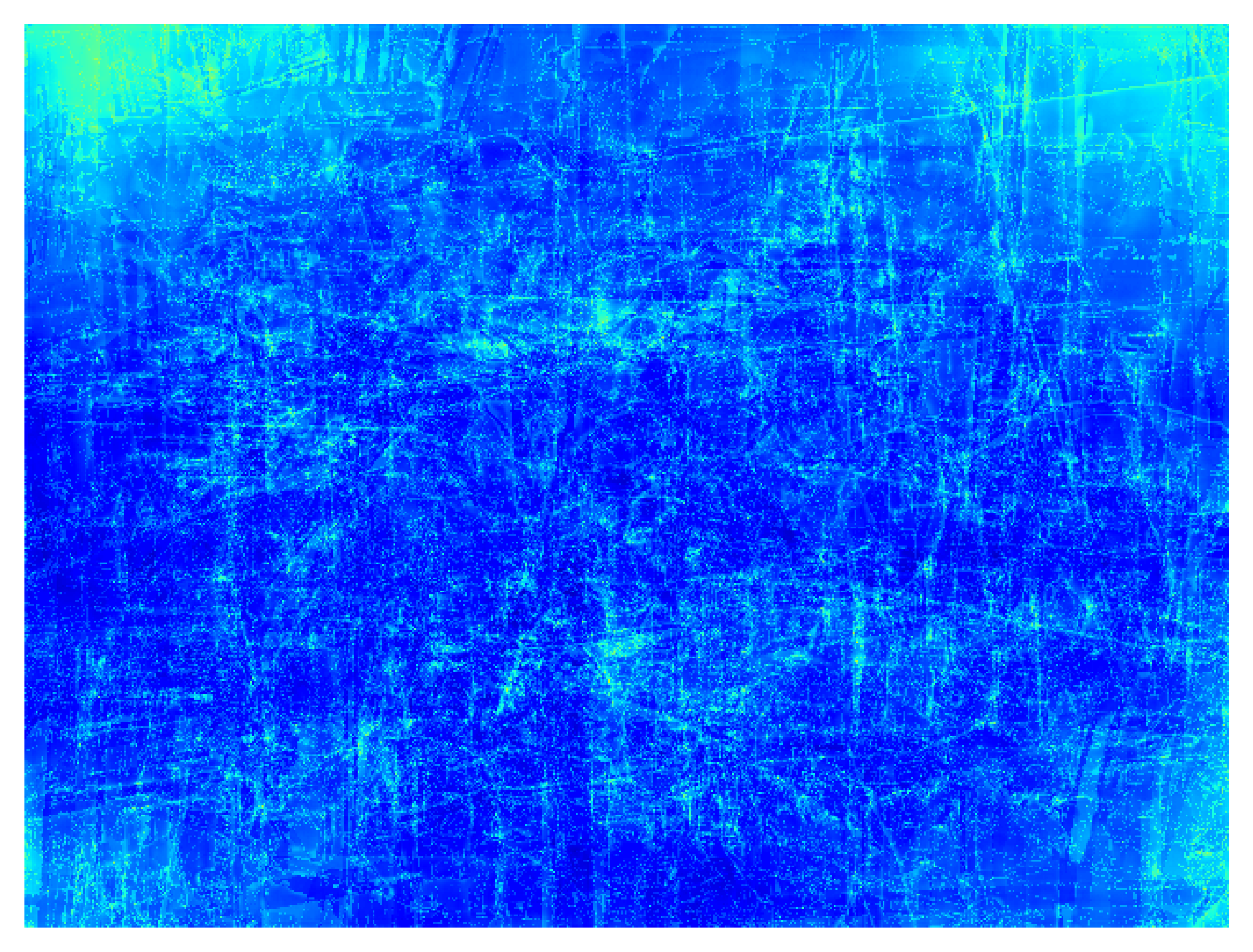}
        \end{tabular}
        }
        \caption{\textbf{Depth estimation error map.} The baseline model incorrectly estimates depth information in the corners of the image. In contrast, the AAT scheme eliminates the effect of optical aberrations in depth estimation and leads to a uniform error map.}
        \label{fig:error_map}
\end{figure}

\highlight{In an optical lens, aberrations typically become more severe as the off-axis angle increases. Therefore, in this section, we evaluate the effectiveness of the AAT scheme for across the FoV. We calculate the relevant error map for the estimated depth maps and average the results, as shown in Fig.~\ref{fig:error_map}. In the figure, the red color represents a large estimation error, while the blue color represents a small estimation error. When trained without the AAT scheme, the network model produces depth maps with significant errors in the corners of the image. These errors are caused by optical aberrations, which affect the determination of the best in-focus frame in a focal stack. In contrast, our proposed AAT scheme takes optical aberrations into account during the training of the DfF  models. As a result, the final estimated depth map is virtually unaffected by off-axis optical aberrations and is more uniform compared to the baseline results.}

% =============================================================
% =============================================================
\subsection{Training Time}

\begin{figure}[!ht]
    \centering
    \includegraphics[width=0.9\linewidth]{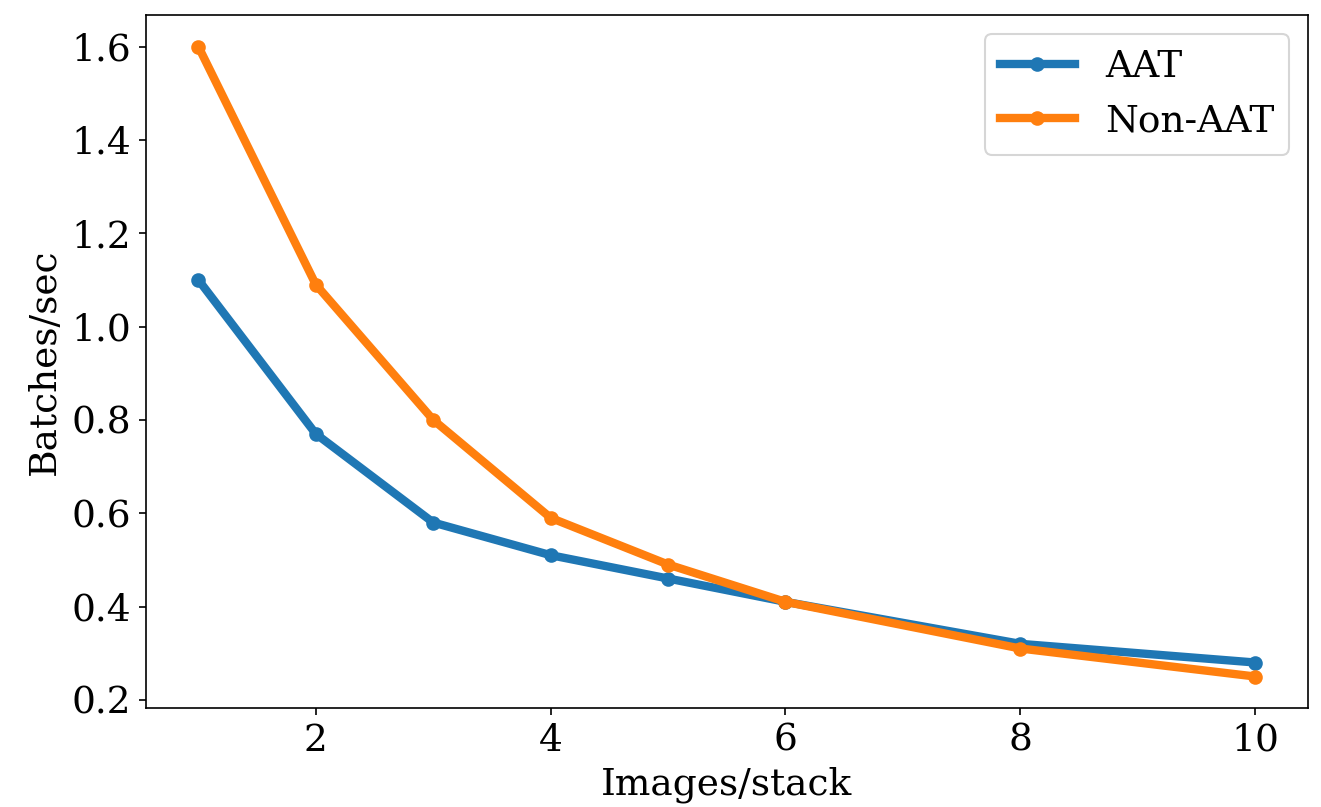}
    \caption{\textbf{Comparison of training speed for different stack sizes.} The AAT scheme has a significantly slower speed when the focal size is small. But when the stack size is large, rendering focal stacks on-the-fly shows no time delay compared to loading focused images from the computer memory. }
    \label{fig:time}
\end{figure}

We also analyze the extra time introduced by the AAT scheme, which results from the image re-rendering process. We use DFVNet to evaluate the training speed with varying numbers of stacks. In Fig.~\ref{fig:time}, we report the average number of batches per second during training. The image resolution is 480$\times$640. For AAT, we render the focus stack for each batch during the training. For comparison, we render all focused images before the experiment and only load them to form focus stacks during the training. The average speed is calculated over 200 batches within a training epoch, and we run the model for 5 epochs to reduce the variance. When the stack size is small ($\leq 4$), the AAT training scheme reduces the training speed. However, in DfF applications, we typically use a stack size larger than 5 in our experiments, and the speed difference is negligible. Particularly, when the number of stacks is large ($\geq 8$), AAT training is faster since it reduces the overhead of image loading.

When the stack size equals 1, we analyze the additional time cost introduced in monocular image processing. The AAT scheme reduces the training speed by approximately $30\%$. However, this is also highly dependent on the network architecture. The DFVNet used for evaluation is a small and simple network; thus, the AAT scheme introduces a significant impact. However, we believe that this extra time can be negligible for more complex and larger network structures.

\section{Discussion}
The experiments conducted in the previous section demonstrate that optical aberrations can degrade the generalizability of pretrained DfF models, but we can improve the results with the proposed AAT scheme. However, there are also some limitations that need to be considered, especially in real-world practice.

Firstly, the impact of optical aberrations is significant only for small defocus distances, as the PSF is dominated by defocus phenomena when object points are far from the focus plane. Therefore, the AAT scheme is more effective in improving depth estimation accuracy for adjacent objects with less significant defocus. However, in sparse scenes, the improvement may not be as significant.

Secondly, the AAT scheme can provide more significant improvements for
lenses with larger aberrations. Current DfF datasets contain fairly low-resolution images, in large part, because the depth sensors used to
generate ground truth data have a limited pixel count. At these low
image resolutions, most commercial-grade lenses only exhibit very
modest aberrations.  However, we believe that the AAT scheme will be
vital for extending the DfF approach to the modern image sensor
resolutions, as well as in the context of end-to-end learned compact
computational imaging lenses~\cite{sun2021end, li2021end, tseng2021neural}.

Thirdly, the focus distance information obtained from the EXIF data of photography cameras is often inaccurate, leading to reduced depth estimation performance. However, modern imaging and display systems typically include Lidar or depth cameras that can provide more accurate focus distance information. By incorporating such information, we believe the depth estimation performance can be further improved, especially in scenarios where multiple cameras are employed.
\section{Conclusion}

In this work, we address the domain gap caused by off-axis optical aberrations, which has been overlooked by most existing works. To this end, we propose an AAT scheme to bridge this gap. Specifically, we develop a network to estimate the PSF of a real camera lens for different positions and focus distances. We then use the estimated PSF to simulate aberrated training images, enabling the network to learn to extract more accurate image features in the presence of optical aberrations. We evaluate the AAT scheme on two DfF networks and demonstrate its generalizability through both simulated and real-world experiments. The experimental results indicate that the AAT scheme improves the generalizability of the DfF models. The DfF models trained with AAT can estimate more accurate depth maps without fine-tuning compared to the baseline. Furthermore, we believe that the AAT scheme is not limited to the DfF task and can be applied to improve the generalizability of other computer vision tasks.

% \lipsum[2-4]

% \begin{figure}[!t]
% \centering
% \framebox[\columnwidth]{\parbox{0.9\columnwidth}{~\\~\\~\\~\\~\\}}
% \caption{Example one-column figure.}
% \end{figure}

% \section{Related Work}
% \lipsum[5-6]

% \begin{figure*}[!t]
% \centering
% \framebox[\textwidth]{\parbox{0.9\textwidth}{~\\~\\~\\~\\~\\~\\~\\}}
% \caption{Example two-column figure.}
% \end{figure*}

% \section{Proposed Method}
% \lipsum[7]

% \subsection{Subsection Heading Here}
% \lipsum[8-9]

% \subsubsection{Subsubsection Heading Here}
% \lipsum[10-12]

% \section{Experimental Results}

% \lipsum[13-17]

% \begin{table}[!t]
% \renewcommand{\arraystretch}{1.3}
% \caption{An Example of a Table}
% \centering
% \begin{tabular}{c||c|c|c}
% \hline
% One & Two & One & Two\\
% \hline\hline
% Three & Four &Three & Four\\
% \hline
% Three & Four &Three & Four\\
% \hline
% Three & Four &Three & Four\\
% \hline
% \end{tabular}
% \end{table}

% \section{Conclusion}
% \lipsum[18]

% Any acknowledgments to only be included in camera ready
\ifpeerreview \else
\section*{Acknowledgments}
This work was supported by the King Abdullah University of Science and Technology (KAUST) individual baseline funding.
\fi

\bibliographystyle{IEEEtran}
\bibliography{ref}

\ifpeerreview \else
%%%% For the camera ready version, please fill out this
%%%% biography. Your camera ready should be within a 12 page limit
%%%% including acknowledgments, references and biography.

% \includegraphics command within an optional argument. (You could
% create your own custom macro containing the \includegraphics command
% to make things simpler here.)
% \begin{IEEEbiography}[{\includegraphics[width=1in,height=1.25in,clip,keepaspectratio]{mshell}}]{Michael Shell}

% \end{IEEEbiography}

% % insert where needed to balance the two columns on the last page with
% %\newpage


% \newpage
% \balance
\vspace{-3cm}
\begin{IEEEbiography}[{\includegraphics[width=1in,height=1.25in,clip,keepaspectratio]{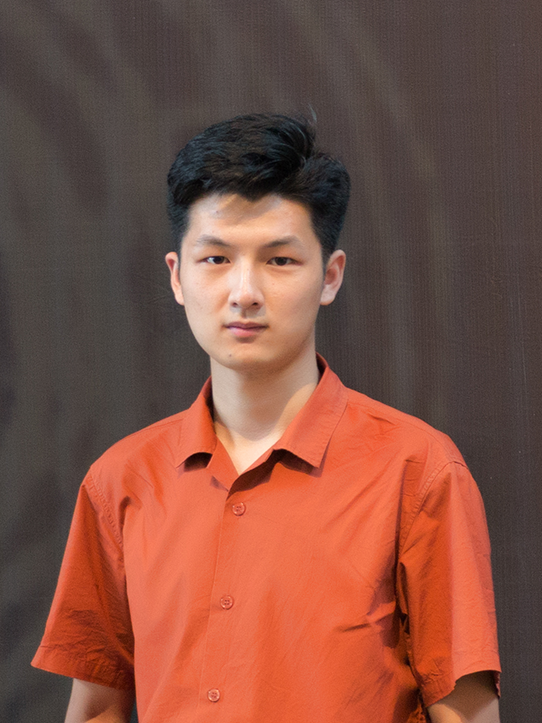}}]{Xinge Yang} received the B.S. degree in physics (major) and computer science (minor) from University of Science and Technology of China in 2020 and the M.Sc. in computer science from King Abdullah University of Science and Technology in 2022. He is currently a Ph.D. student at King Abdullah University of Science and Technology. His research interests include computational imaging, computational lens design, and differentiable optical simulation.
\end{IEEEbiography}

\vspace{-3cm}
\begin{IEEEbiography}[{\includegraphics[width=1in,height=1.25in,clip,keepaspectratio]{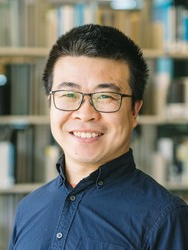}}]{Qiang Fu} received the B.S. degree in Mechanical Engineering from University of Science and Technology of China, Hefei, China in 2007 and the Ph.D. degree in Optical Engineering from University of Chinese Academy of Sciences, Beijing, China in 2012. He is currently a Research Scientist at the Visual Computing Center, KAUST. Previously, he was a Research Associate Professor at ShanghaiTech University (2016-2017), Postdoctoral Research Fellow at KAUST (2014-2016), Assistant Research Fellow at Academy of Opto-Electronics, Chinese Academy of Sciences (2012-2014). His research interests lie in the multidisciplinary research areas of computational imaging, including optical system design, diffractive optics, micro/nano-fabrication, hyperspectral imaging, polarization imaging, high dynamic range imaging, wavefront sensing etc.
\end{IEEEbiography}

\vspace{-3cm}
\begin{IEEEbiography}[{\includegraphics[width=1.0in,height=1.0in,clip,keepaspectratio]{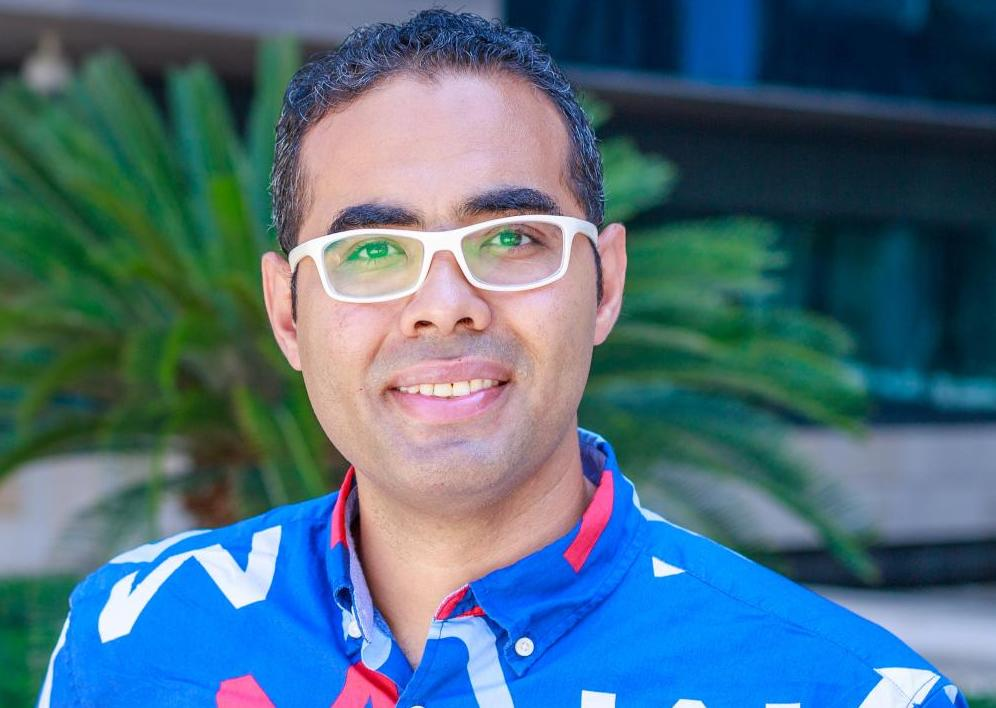}}]{Mohamed Elhoseiny}
is an assistant professor of Computer Science at KAUST. Researcher at Facebook Research. Previously, he was a visiting Faculty at Stanford Computer Science department (2019-2020), Visiting Faculty at Baidu Research Silicon Valley Lab (2019), and Postdoc researcher at Facebook AI Research (2016-2019). Dr. Elhoseiny did his Ph.D. in 2016 at Rutgers University during which he spent time at Adobe Research (2015-2016) for more than a year and at SRI International in 2014 (a best intern award). Dr. Elhoseiny received an NSF Fellowship in 2014 for the Write-a-Classifier project (ICCV13) and the Doctoral Consortium award at CVPR 2016. Dr. Elhoseiny was also a Stanford Igniter as a venture idea generator; Stanford Ignite is an entrepreneurship program at Stanford Graduate School of Business. His primary research interest is in computer vision and especially in efficient multimodal learning with limited data in areas like zero/few-shot learning,  Vision \& Language, language guided visual perception. He is also interested in Affective AI and especially to produce novel art and fashion with AI. His creative AI work was featured in MIT Tech Review,  New Scientist Magazine, and HBO Silicon Valley.
\end{IEEEbiography}

\newpage

\begin{IEEEbiography}[{\includegraphics[width=1in,height=1.25in,clip,keepaspectratio]{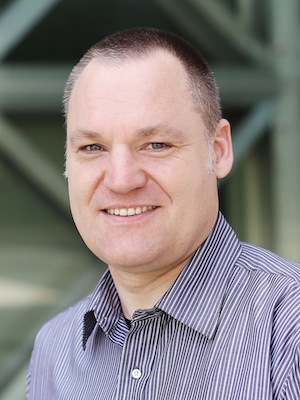}}]{Wolfgang Heidrich}
  (Fellow, IEEE) is a Professor of Computer Science
  and Electrical and Computer Engineering in the KAUST Visual
  Computing Center, for which he also served as director from 2014 to
  2021. Heidrich joined KAUST in 2014, after 13 years as a faculty
  member at the University of British Columbia. He received his PhD in
  from the University of Erlangen in 1999, and then worked as a
  Research Associate in the Computer Graphics Group of the
  Max-Planck-Institute for Computer Science in Saarbrucken, Germany,
  before joining UBC in 2000. Heidrich’s research interests lie at the
  intersection of imaging, optics, computer vision, computer graphics,
  and inverse problems. His more recent interest is in computational
  imaging, focusing on hardware-software co-design of the next
  generation of imaging systems, with applications such as
  High-Dynamic Range imaging, compact computational cameras,
  hyperspectral cameras, to name just a few. His work on HDR displays
  serves as the basis for many modern-day HDR consumer devices today.
  Heidrich is a Fellow of the IEEE, AAIA, and Eurographics, and the
  recipient of a Humboldt Research Award as well as the ACM SIGGRAPH
  Computer Graphics Achievement Award.
\end{IEEEbiography}

%\vfill

\fi

\end{document}

\IEEEtitleabstractindextext{%
\begin{abstract}
Computer vision methods for depth estimation usually use simple camera models with idealized optics. For modern machine learning approaches, this creates an issue when attempting to train deep networks with simulated data, especially for focus-sensitive tasks like Depth-from-Focus. In this work, we investigate the domain gap caused by off-axis aberrations that will affect the decision of the best-focused frame in a focal stack. We then explore bridging this domain gap through aberration-aware training (AAT). Our approach involves a lightweight network that models lens aberrations at different positions and focus distances, which is then integrated into the conventional network training pipeline. We evaluate the generality of network models on both synthetic and real-world data. The experimental results demonstrate that the proposed AAT scheme can improve depth estimation accuracy without fine-tuning the model for different datasets. The code and models will be made publicly available.
\end{abstract}

\begin{IEEEkeywords} % Enter keywords here
Depth from Focus, Optical Aberration, Ray Tracing, Point Spread Function
\end{IEEEkeywords}
}

% Make Title
\ifpeerreview
\linenumbers \linenumbersep 15pt\relax 
\author{Paper ID \paperID\IEEEcompsocitemizethanks{\IEEEcompsocthanksitem This paper is under review for ICCP 2023 and the PAMI special issue on computational photography. Do not distribute.}}
\markboth{Anonymous ICCP 2023 submission ID \paperID}%
{}
\fi
\maketitle
\thispagestyle{empty}

% The first section title should be wrapped inside a \IEEEraisesectionheading as follows.
% \IEEEraisesectionheading{
%   \section{Introduction}\label{sec:introduction}
% }
% The very first letter of the paper is a 2 line initial drop letter
% followed by the rest of the first word in caps.
% \IEEEPARstart{T}{his} demo file is intended to serve as a ``starter
% file'' for ICCP 2023 submissions produced under
% \LaTeX~\cite{kopka-latex} using IEEEtran.cls version 1.8b and later.

% For submissions for review, the paper uses the lineno package which
% might require you to compile under latex twice to get the line numbers
% to align correctly.

% \input{1_intro.tex}
% \input{2_related_works.tex}
% \input{3_methods.tex}
% \input{4_1_psfnet_experiments.tex}
% \input{4_2_dff_experiments.tex}
% \input{5_discussion.tex}
% \input{6_conclusions.tex}
\section{Dataset}

\subsection{Image Dataset}

Our experiments involve two types of focal stacks: simulated focal stacks generated using RGBD images and real-world captured focal stacks. Simulated focal stacks comprise both synthetic scenes and real-world scenes, while real-world captured focal stacks consist of both outdoor scenes and indoor scenes. To facilitate aberration-aware training, we require access to lens data, focus distances, focal stacks, and depth maps. Unfortunately, no such dataset is currently available, so we created our own training data based on simulated focal stacks, which allowed us to leverage known lens models and select appropriate focus distances. For testing, we used both simulated focal stacks and real-world captured focal stacks.

\begin{figure*}[!ht]
    \centering
    \setlength{\tabcolsep}{1pt}
    \resizebox{0.8\textwidth}{!}{
    \begin{tabular}{cc}
         \includegraphics[width=0.45\textwidth]{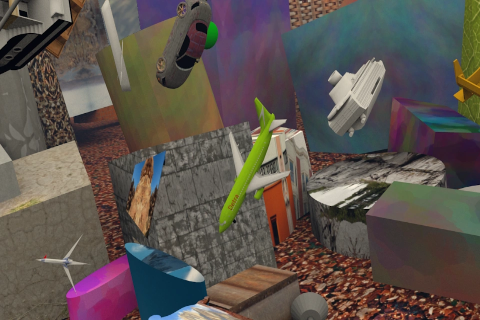}
         &
         \includegraphics[width=0.45\textwidth]{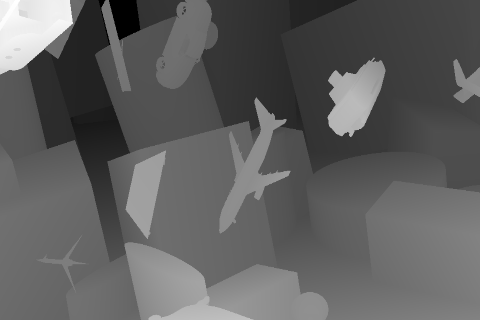}
         \\
         (a) All-in-focus RGB image
         &
         (b) Ground-truth depth map
         \\
         \includegraphics[width=0.45\textwidth]{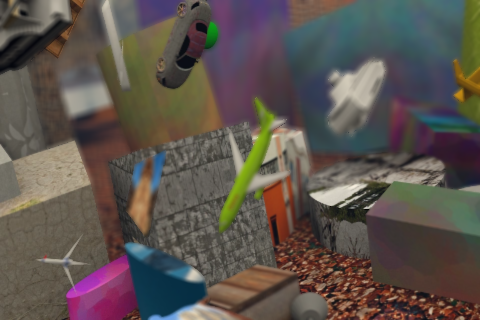}
         &
         \includegraphics[width=0.45\textwidth]{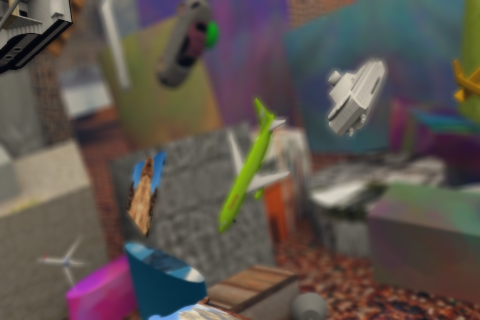}
         \\
         (c) Focused image example 1
         &
         (d) Focused image example 2
    \end{tabular}
    }
    \caption{Simulated focal stacks with RGBD images.}
    \label{fig_supp:simulated}
\end{figure*}

To generate simulated focal stacks using RGBD data, we select the FlyingThings3D~\cite{mayer2016large, wang2021bridging}, Middlebury2014~\cite{scharstein2014high}, and Matterport3D~\cite{zhang2018deep} datasets, which provide RGBD images. To simulate the focal stacks, we use the normalized pixel coordinates, focus distance, and depth map to calculate the point spread function (PSF) for each pixel, and then perform a convolution to generate focused images. The focus distances are chosen linearly within the minimum and maximum depth range of each image to mimic the focal swapping process. In addition to regular data augmentation techniques, such as random flips and color jittering, we also introduce a random perturbation to each focus distance to enhance the variety of our simulated focal stacks. An example of our simulated focal stacks can be seen in Fig.~\ref{fig_supp:simulated}.

\begin{figure*}[!ht]
    \centering
    \setlength{\tabcolsep}{1pt}
    \resizebox{0.8\textwidth}{!}{
    \begin{tabular}{cc}
         \includegraphics[width=0.45\textwidth]{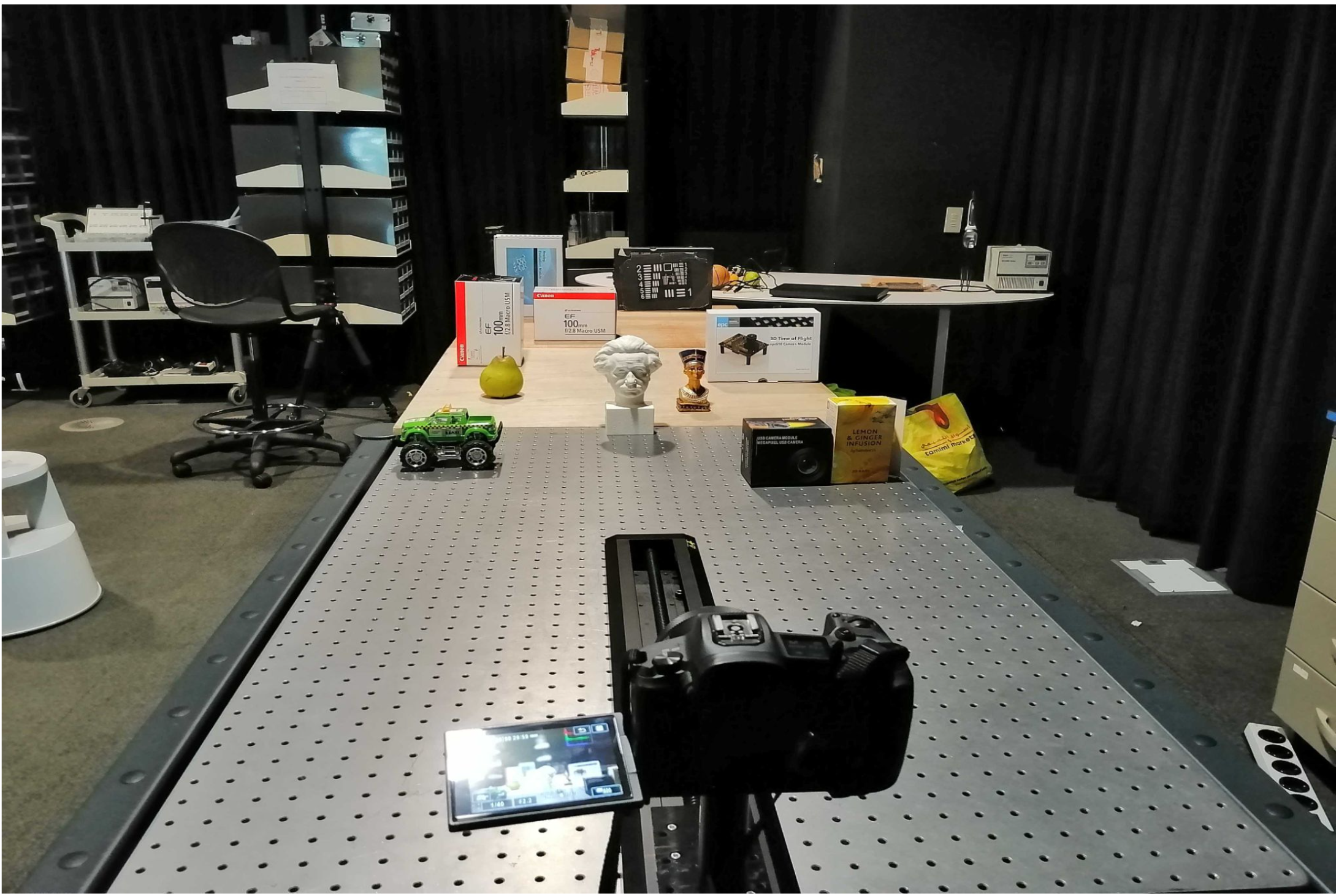}
         &
         \includegraphics[width=0.45\textwidth]{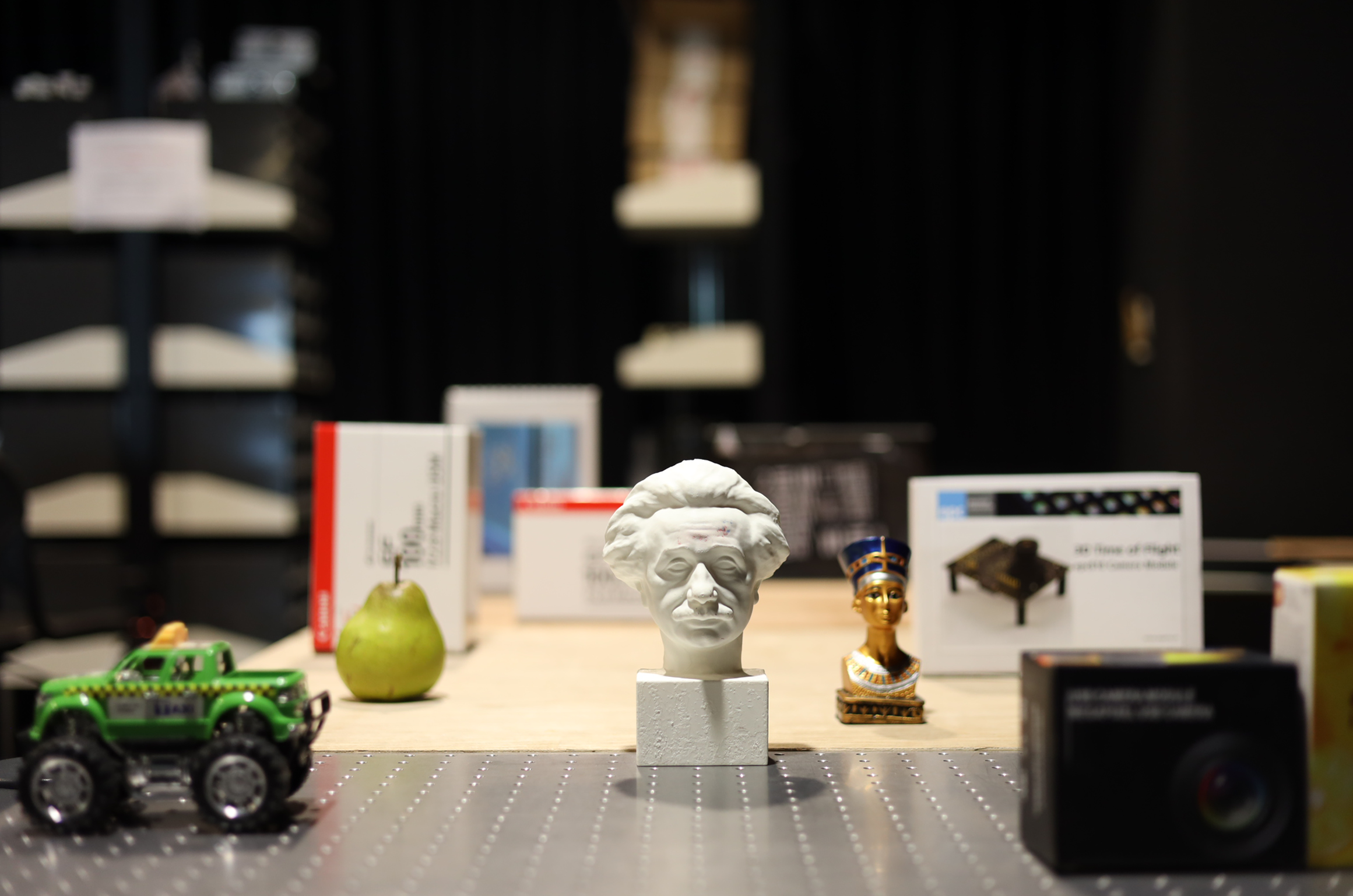}
         \\
         (a) Experimental setup
         &
         (b) Focused image example
         \\
         \includegraphics[width=0.45\textwidth]{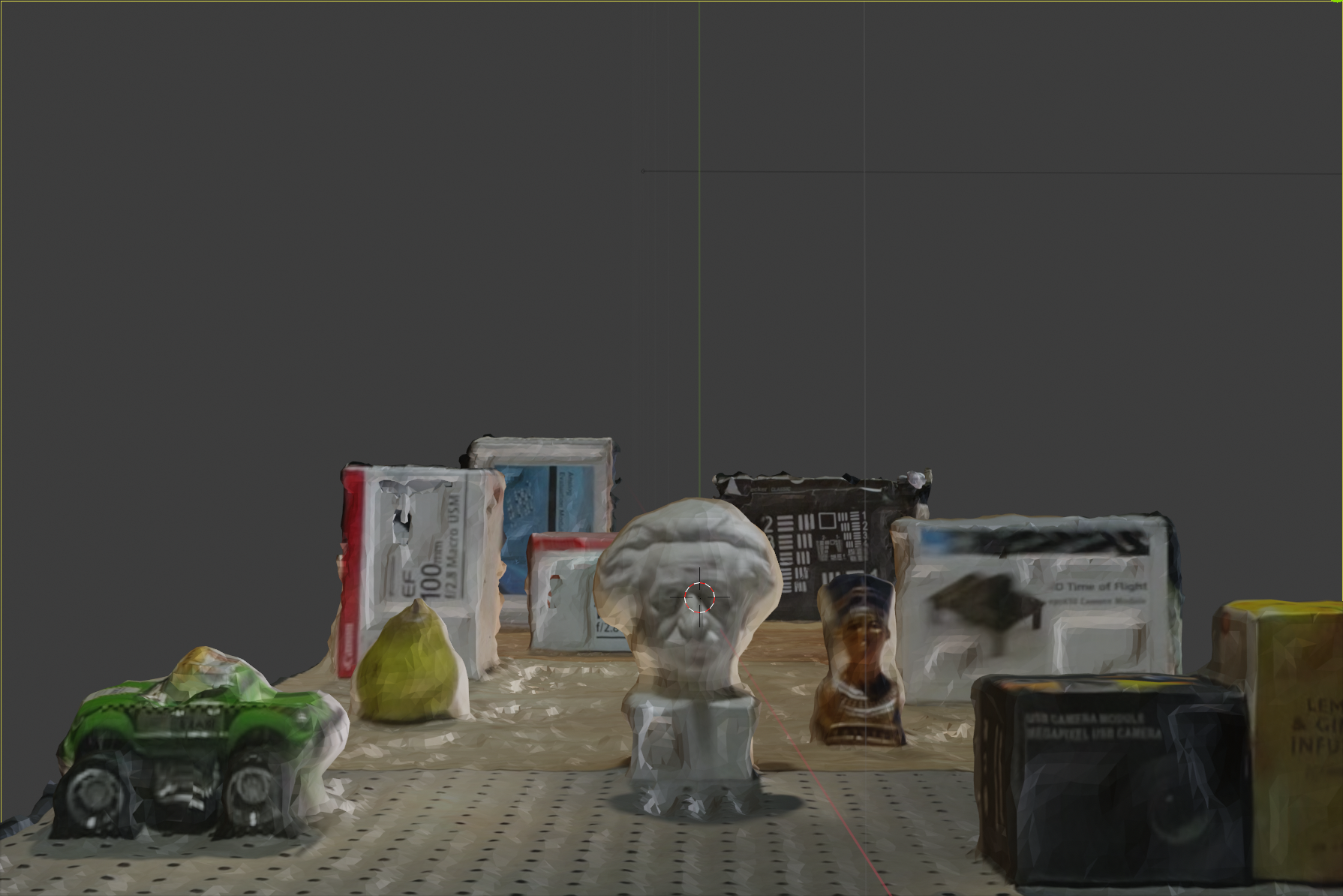}
         &
         \includegraphics[width=0.45\textwidth]{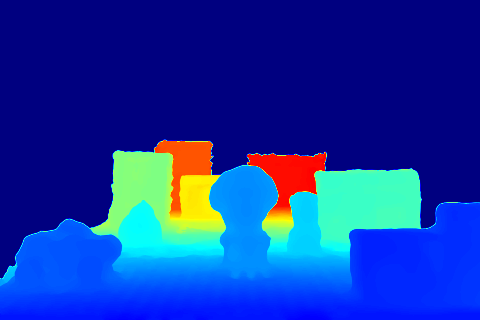}
         \\
         (c) LiDAR scan result
         &
         (d) Rendered depth map
    \end{tabular}
    }
    \caption{Real-world indoor scene. (a) We set up an indoor scene with objects placed at different positions. (b) We focus the camera on different objects and capture focal stacks. (c) We use the LiDAR sensor on iPhone 14 pro/iPad pro to scan the 3D scene and load the scanned data into Blender. We adjust the camera position and view angle to align with the focused image. (d) We render the depth map with Blender and use it as the ground truth for quantitative evaluation.}
    \label{fig:indoor}
\end{figure*}

To capture our real-world captured focal stacks, we position the camera and focus it on different objects while measuring the distance from the focused objects to the camera (for the indoor scene), or reading the focus distances from the EXIF metadata of the photos (for the outdoor scenes). Our real-world focal stacks comprise 24 outdoor scenes and 1 indoor scene. Fig.~\ref{fig:indoor} depicts the experimental setup for the indoor scene. We use the LiDAR sensor on an iPhone 14 pro to scan the 3D scene and load the scanned data into Blender software. We then adjust the camera position and view angle to align with the focused images. Once aligned, we render the depth map with Blender and use it as the ground truth for quantitative evaluation. Due to the breathing effect, the focused images are not well-aligned. We load them into Photoshop, align them, and then output the aligned focal stack.

The resolution ratio of training and testing images is determined by the camera sensor resolution, and any resizing operation can alter the aberration property of the lens. However, downsampling is feasible due to the physical properties of the PSF. Although optical aberrations are more pronounced at higher image resolutions, network training processes cannot handle the extremely high-resolution images produced by commercial cameras with megapixel sensors. Therefore, we find a balance between performance and memory consumption that allows us to observe aberrations in the image without using excessive memory during network training. In our experiments, we used two different image resolutions: 480 $\times$ 640 for the 50mm F/2.8 lens and 640 $\times$ 960 for the Canon RF 50mm F/1.8 lens. As shown in Fig.3 (main paper) and Fig.~\ref{fig:psf_50mm}, the off-axis aberrations can be easily seen at the resolutions we use.

% \begin{figure*}[!ht]
%     \centering
%     \setlength{\tabcolsep}{1pt}

%     \includegraphics[width=\textwidth]{supp_fig/indoor_scene/fig.png}

%     \caption{Lens setup}
%     \label{fig:indoor}
% \end{figure*}

\subsection{Lens Data}

\begin{figure*}[!ht]
    \centering
    \setlength{\tabcolsep}{1pt}
    \resizebox{\textwidth}{!}{
    \begin{tabular}{ccc}
         \includegraphics[width=3cm]{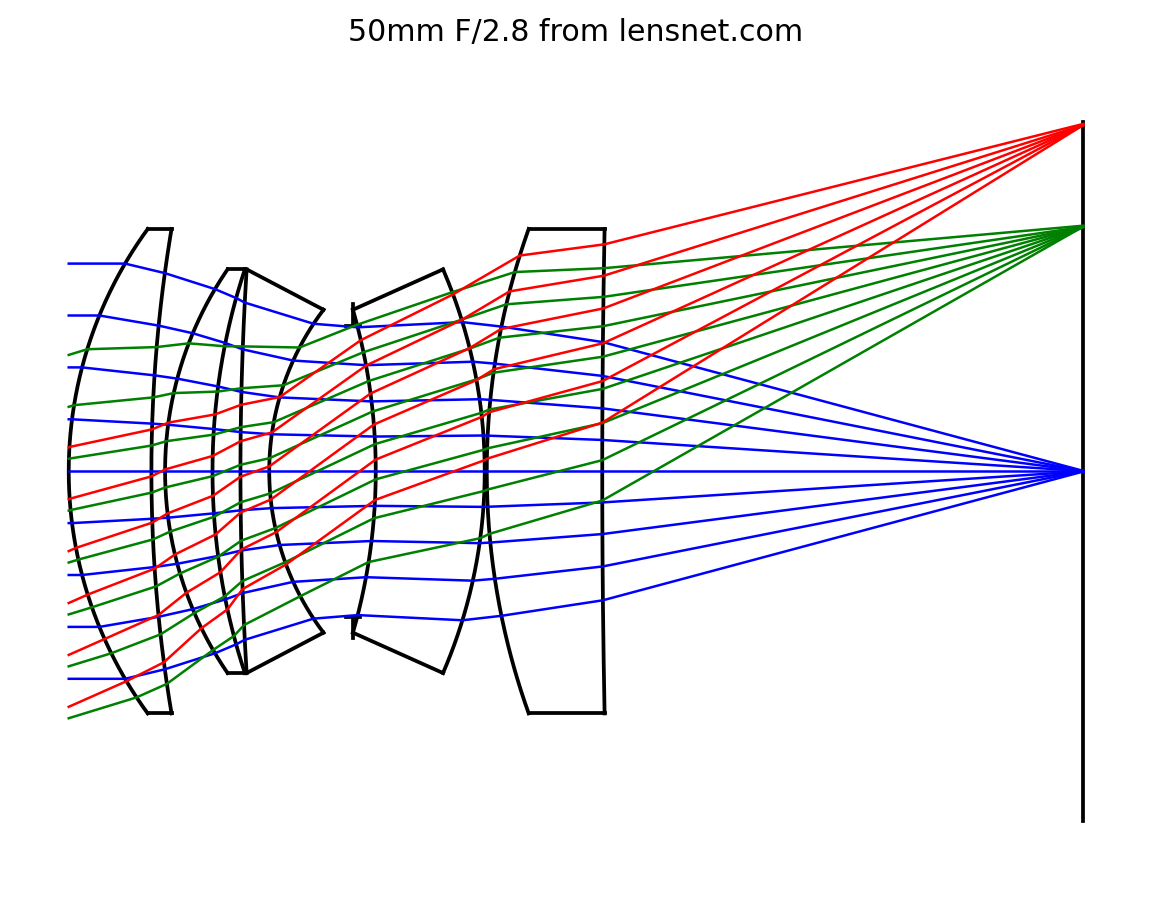}
         &
         \includegraphics[width=3cm]{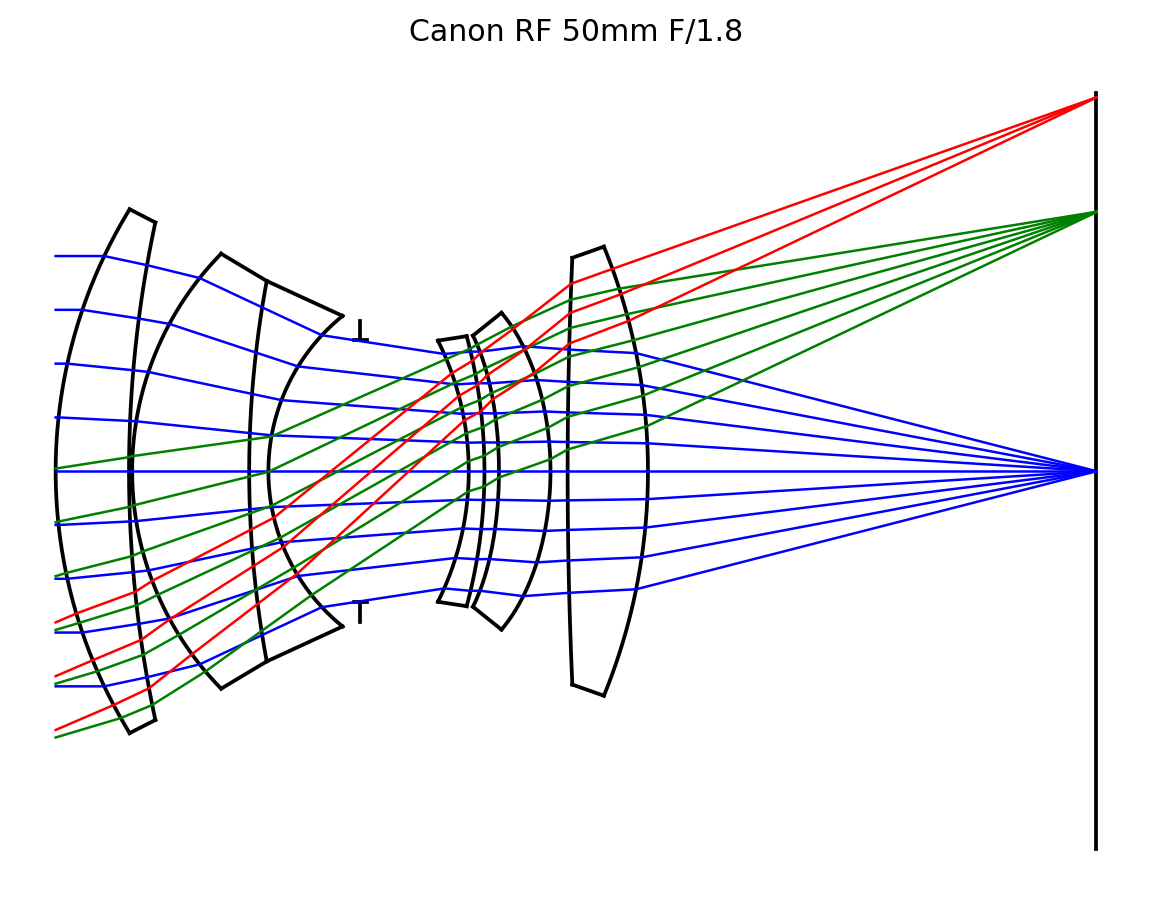}
    \end{tabular}
    }
    \caption{Lens setup. Left: a 50mm F/2.8 lens from \href{http://lensnet.herokuapp.com/}{lensnet.com}~\cite{cote2021deep}. Right: the Canon RF 50mm F/1.8 lens.}
    \label{fig:lens}
\end{figure*}

\begin{figure*}[htp]
    \centering
    \includegraphics[width=\textwidth]{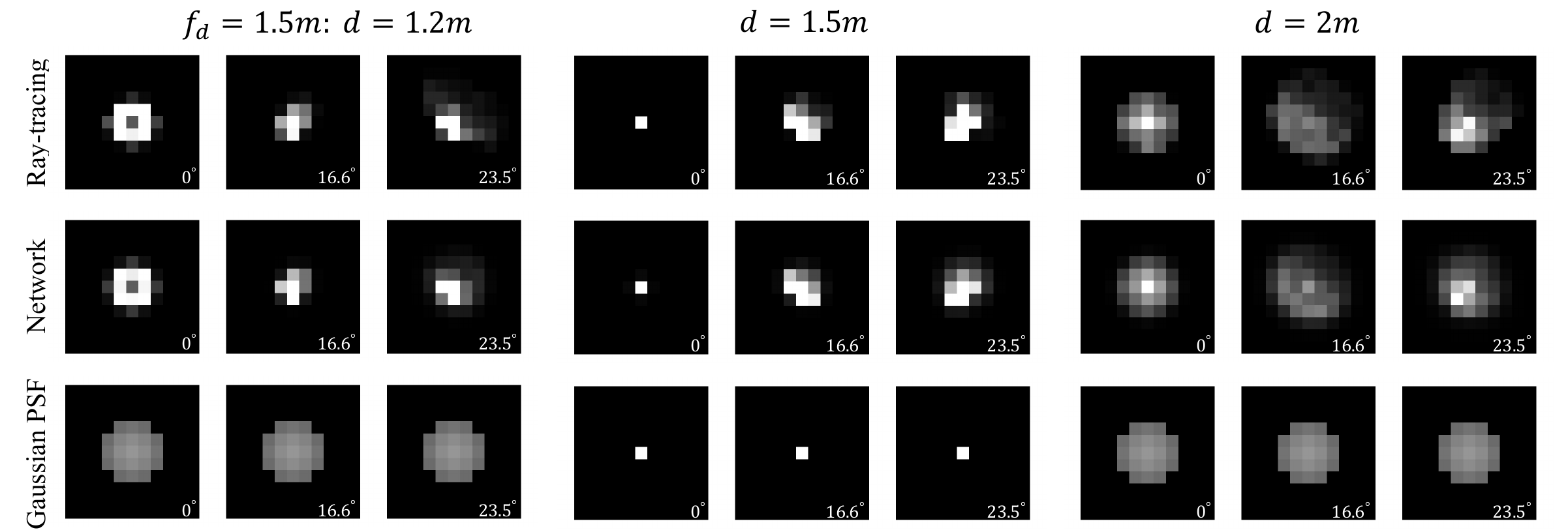}
    \caption{The PSF evaluation results for the 50mm F/2.8 lens. We focus the lens to a distance of 1.5m and evaluate the PSF of the three methods at three depths and three view angles. Our proposed PSF network produces a PSF that is similar to the ground truth (ray tracing). In contrast, the Gaussian PSF exhibits significant differences, particularly at large view angles. The PSF of a real lens varies with different view angles due to the presence of off-axis optical aberrations, while the Gaussian PSF model neglects this aberration.}
    \label{fig:psf_50mm}
\end{figure*}

Our experiments involved two lenses: the 50mm F/2.8 lens, generated using \href{http://lensnet.herokuapp.com/}{lensnet.com~\cite{cote2021deep}}, and the Canon RF 50mm F/1.8 lens, a commercially available lens. The lens data for both lenses is presented in Table.~\ref{lens:lensnet} and Table.\ref{lens:canon}, respectively. The 2D lens structures with ray paths are displayed in Fig.~\ref{fig:lens}. Notably, the 50mm F/2.8 lens exhibits poorly corrected optical aberrations, thus the performance improvement with the AAT scheme is more obvious compared to the Canon lens which has well-corrected optical aberrations.

\begin{table*}[ht]
    \centering
    % \scriptsize
    \caption{Lens data for the 50mm F/2.8 lens used in the paper. The lens is generated by~\cite{cote2021deep}.}
    \resizebox{0.98\textwidth}{!}{
    \begin{tabular}{lcccccccccc}
    \hline
    Surface & Radius [mm] & Thickness [mm] & Material (n/V) & Semi-diameter [mm] & Conic & $\alpha_{4}$ & $\alpha_{6}$ & $\alpha_{8}$ & $\alpha_{10}$ & $\alpha_{12}$ \\
    \hline
    1 (Sphere) & 25.445 & 5.120 & 1.7290/54.494 & 15.00 & 0 & 0 & 0 & 0 & 0 & 0  \\
    2 (Sphere) & 90.909 & 0.847 &  & 15.00 & 0 & 0 & 0 & 0 & 0 & 0  \\
    3 (Sphere) & 22.124 & 2.937 & 1.6517/58.5020 & 12.50 & 0 & 0 & 0 & 0 & 0 & 0  \\
    4 (Sphere) & 40.000 & 1.731 &  & 12.50 & 0 & 0 & 0 & 0 & 0 & 0  \\
    5 (Sphere) & 204.082 & 1.782 & 1.6990/30.1789 & 12.50 & 0 & 0 & 0 & 0 & 0 & 0  \\
    6 (Sphere) & 16.556 & 5.183 &  & 10.00 & 0 & 0 & 0 & 0 & 0 & 0 \\
    7 (Aper) &  & 1.414 &  & 9.00 & 0 & 0 & 0 & 0 & 0 & 0 \\
    8 (Sphere) & -36.101 & 6.749 & 1.6204/60.3100 & 10.00 & 0 & 0 & 0 & 0 & 0 & 0 \\
    9 (Sphere) & -31.546 & 0.127 &  & 12.50 & 0 & 0 & 0 & 0 & 0 & 0  \\
    10 (Sphere) & 44.643 & 7.151 & 1.7551/52.2945 & 15.00 & 0 & 0 & 0 & 0 & 0 & 0 \\
    11 (Sphere) & 769.231 & 29.792 & & 15.00 & 0 & 0 & 0 & 0 & 0 & 0 \\
    Sensor &  &  &  & 21.63 &  & & & & & \\
    \hline
    \end{tabular}
    }
    \label{lens:lensnet}
\end{table*}

\begin{table*}[ht]
    \centering
    % \scriptsize
    \caption{Lens data for the Canon RF 50mm F/1.8 lens used in the paper.}
    \resizebox{0.98\textwidth}{!}{
    \begin{tabular}{lcccccccccc}
    \hline
    Surface & Radius [mm] & Thickness [mm] & Material (n/V) & Semi-diameter [mm] & Conic & $\alpha_{4}$ & $\alpha_{6}$ & $\alpha_{8}$ & $\alpha_{10}$ & $\alpha_{12}$ \\
    \hline
    1 (Sphere) & 28.621 & 4.20 & N-LASF41 & 15.00 & 0 & 0 & 0 & 0 & 0 & 0  \\
    2 (Sphere) & 68.136 & 0.18 &  & 14.24 & 0 & 0 & 0 & 0 & 0 & 0  \\
    3 (Sphere) & 17.772 & 6.70 & N-LASF43 & 12.45 & 0 & 0 & 0 & 0 & 0 & 0  \\
    4 (Sphere) & 59.525 & 1.10 & SF6 & 10.89 & 0 & 0 & 0 & 0 & 0 & 0  \\
    5 (Sphere) & 11.427 & 5.27 &  & 8.89 & 0 & 0 & 0 & 0 & 0 & 0  \\
    6 (Aper) &  & 6.20 &  & 7.50 & 0 & 0 & 0 & 0 & 0 & 0 \\
    7 (Sphere) & -16.726 & 0.90 & SF5 & 7.48 & 0 & 0 & 0 & 0 & 0 & 0 \\
    8 (Sphere) & -29.829 & 0.83 &  & 7.73 & 0 & 0 & 0 & 0 & 0 & 0  \\
    9 (ASphere) & -25.000 & 2.95 & K5G20 & 7.76 & 0 & -4.1203e-5 & -2.9002e-7 & -4.6712e-9 & 7.9065e-11 & -9.2847e-13 \\
    10 (ASphere) & -18.373 & 0.98 & & 9.07 & 0 & -2.4162e-5 & -3.2915e-7 & 1.9110e-10 & -9.2859e-13 & -2.2919e-13 \\
    11 (Sphere) & 280.004 & 4.60 & N-LAK10 & 12.22 & 0 & 0 & 0 & 0 & 0 & 0 \\
    12 (Sphere) & -34.002 & 25.67 & & 12.86 & 0 & 0 & 0 & 0 & 0 & 0 \\
    Sensor &  &  &  & 21.63 &  & & & & & \\
    \hline
    \end{tabular}
    }
    \label{lens:canon}
\end{table*}

\section{PSF Network Training and Evaluation}

In order to effectively train the PSF network, we needed to sample a sufficient amount of input data to account for the complex optical properties, particularly since the PSF varies significantly with the object position and focus distance. When the focus distance is short, even small changes in distance can produce large variations in the PSF. Additionally, for depths close to the focus distance, the PSF undergoes rapid changes, requiring the sampling of numerous depths in the vicinity of the focus distance.

In our experiments, we utilize the principle of importance sampling to optimize the sampling strategy. Specifically, we sample more focus distances from the near range and more depths in the vicinity of the focus distance. For the sampling of normalized off-axis $x$ and $y$ coordinates, we use a uniform distribution.

For each training iteration, we generate 256 input data with different object positions but the same focus distance. We use the same focus distance within each iteration because changing the focus distance requires adjusting the sensor position of the lens model. However, for different iterations, we use different focus distances. We train the PSF network for 100,000 iterations, which we deem sufficient to cover most situations. Once trained, we freeze the parameters and use the PSF network in subsequent depth-from-focus training to generate photorealistic images.

There are other methods for representing the PSF using neural networks. For example, Tseng et al.~\cite{tseng2021differentiable} propose an architecture with an MLP encoder and a CNN decoder to estimate the PSF grid at given depths. We modify the input to exclude the optics parameters and include focus distance parameters, and train the network for comparison. Our pure-MLP network outperforms the other network in terms of PSF representation accuracy. Moreover, our PSF network is better suited for the DfF problem, as different image pixels have different depths, whereas the other PSF network tends to estimate the PSF grid for an entire depth plane. Thus, we can use our PSF network to simultaneously estimate PSFs for points with different depths, while the other network is harder to implement for this purpose.

\section{Depth-from-Focus Network Training and Evaluation}

\subsection{Local PSF Convolution}

Existing methods usually use the same PSF kernel for convolving the entire image. However, in our experiments, we use different PSF kernels for each pixel. This local convolution operation does not have a specific function, thus we provide our implementation in PyTorch, as follows:

\begin{lstlisting}
def local_psf_render(input, psf, kernel_size=11):
""" Blurs image with dynamic Gaussian blur.

Args:
    input (Tensor): The image to be blurred (N, C, H, W).
    psf (Tensor): Per pixel local PSFs (1, H, W, ks, ks)
    kernel_size (int): Size of the PSFs. Defaults to 11.

Returns:
    output (Tensor): Rendered image (N, C, H, W)
"""

if len(input.shape) < 4:
    input = input.unsqueeze(0)

b,c,h,w = input.shape
pad = int((kernel_size-1)/2)

# 1. pad the input with replicated values
inp_pad = torch.nn.functional.pad(input, pad=(pad,pad,pad,pad), mode='replicate')
# 2. Create a Tensor of varying Gaussian Kernel
kernels = psf.reshape(-1, kernel_size, kernel_size)
kernels_rgb = torch.stack(c*[kernels], 1)
# 3. Unfold input
inp_unf = torch.nn.functional.unfold(inp_pad, (kernel_size,kernel_size))   
# 4. Multiply kernel with unfolded
x1 = inp_unf.view(b,c,-1,h*w)
x2 = kernels_rgb.view(b, h*w, c, -1).permute(0, 2, 3, 1)
y = (x1*x2).sum(2)
# 5. Fold and return
return torch.nn.functional.fold(y,(h,w),(1,1))
\end{lstlisting}

For high-resolution images, we may encounter memory issues. One solution is to perform local PSF convolution on image patches, which can be implemented as follows:

\begin{lstlisting}
def local_psf_render_high_res(input, psf, patch_size=[320, 480], kernel_size=11):
    B, C, H, W = input.shape
    img_render = torch.zeros_like(input)
    for pi in range(int(np.ceil(H/patch_size[0]))):
        for pj in range(int(np.ceil(W/patch_size[1]))):
            low_i = pi * patch_size[0]
            up_i = min((pi+1)*patch_size[0], H)
            low_j = pj * patch_size[1]
            up_j =  min((pj+1)*patch_size[1], W)

            img_patch = input[:, :, low_i:up_i, low_j:up_j]
            psf_patch = psf[:, low_i:up_i, low_j:up_j, :, :]

            img_render[:, :, low_i:up_i, low_j:up_j] = local_psf_render(img_patch, psf_patch, kernel_size=kernel_size)
    
    return img_render
\end{lstlisting}

\subsection{DfF Training and Evaluation}

Simulating focal stacks from RGBD datasets allows us to use desired lens models and flexible focus distances. During the depth-from-focus (DfF) training, we follow the original training procedure and hyperparameter settings as reported in the original papers for the AiFNet~\cite{wang2021bridging} and the DFVNet~\cite{yang2022deep}. We find that stacking the focused images in order of focus distance leads to better training results than stacking them randomly. Therefore, we form both training and testing focal stacks following this principle.

\begin{figure*}[!ht]
    \centering
    \setlength{\tabcolsep}{1pt}
    \resizebox{\textwidth}{!}{
    \begin{tabular}{ccccc}
         GT & Ours (AiFNet) & Baseline (AiFNet) & Ours (DFVNet) & Baseline (DFVNet)
         \\
         \includegraphics[width=3cm]{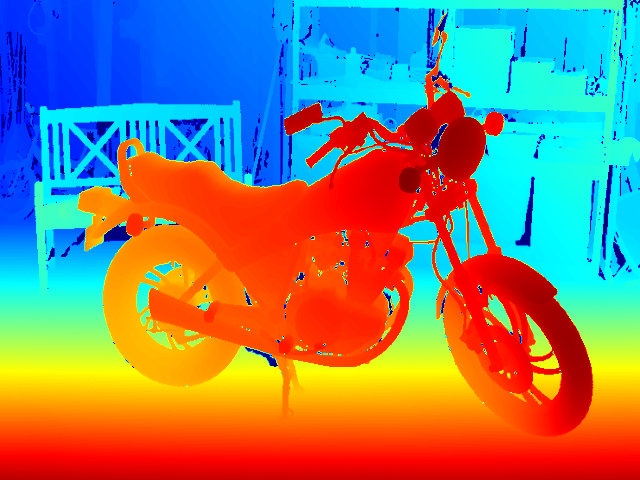}
         &
         \includegraphics[width=3cm]{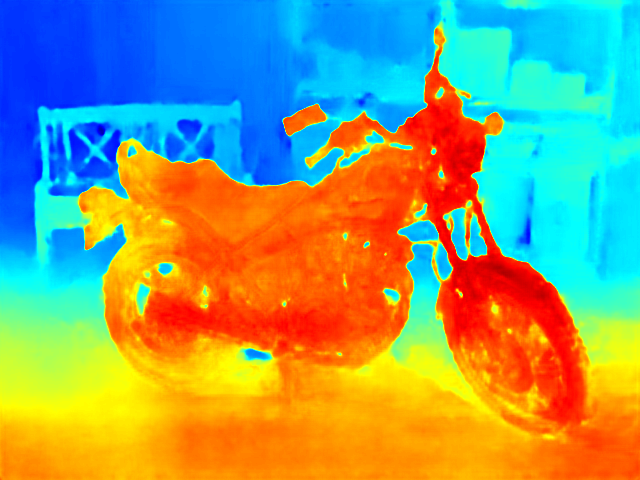}
         & 
         \includegraphics[width=3cm]{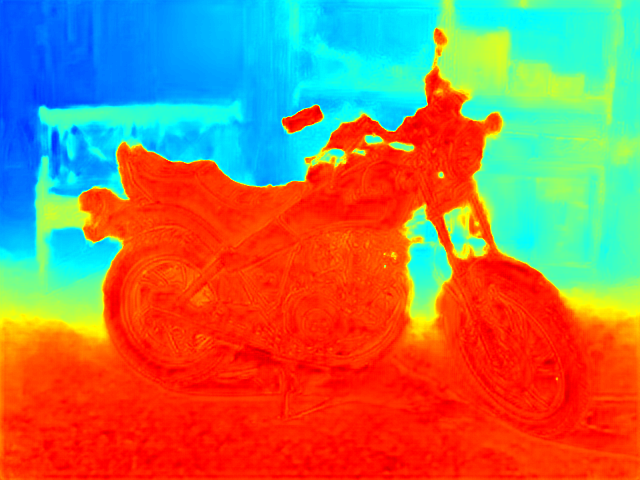}
         &
         \includegraphics[width=3cm]{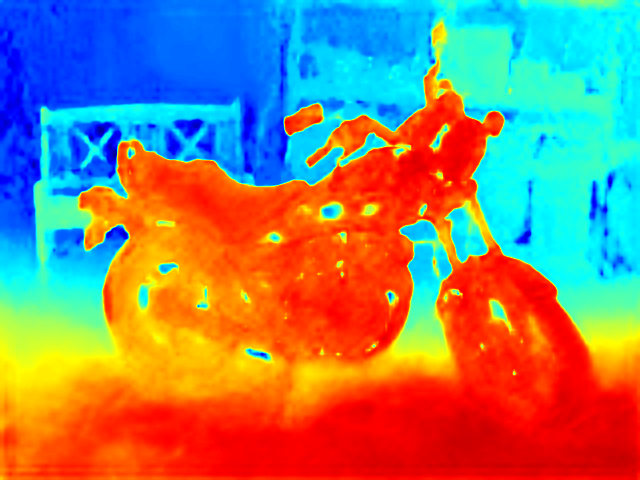}
         &
         \includegraphics[width=3cm]{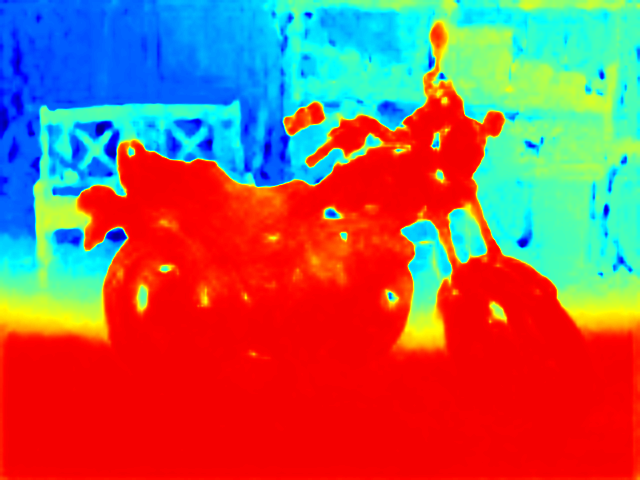}
         \\
         \includegraphics[width=3cm]{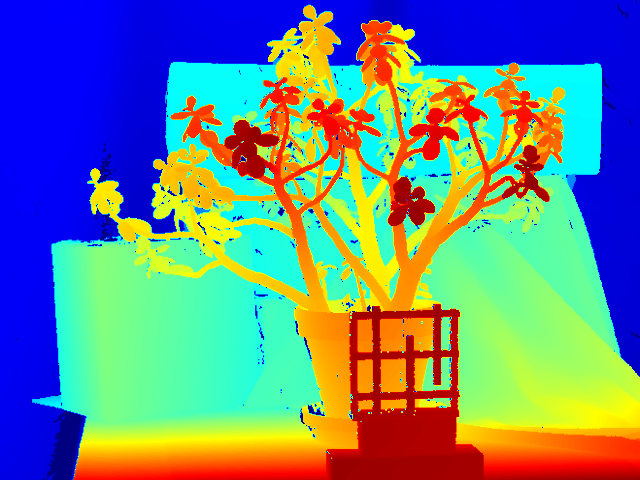}
         &
         \includegraphics[width=3cm]{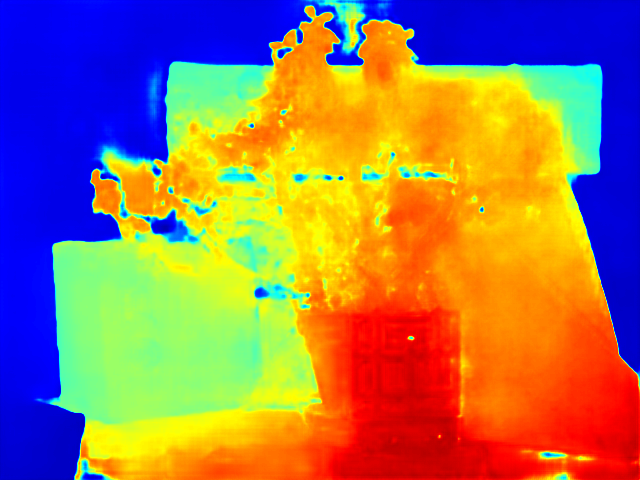}
         & 
         \includegraphics[width=3cm]{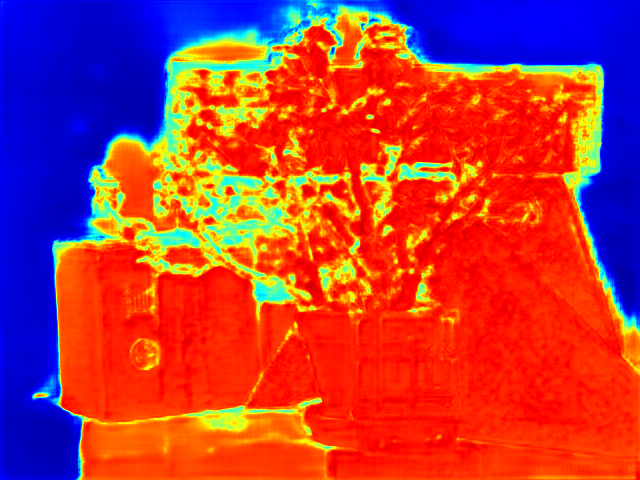}
         &
         \includegraphics[width=3cm]{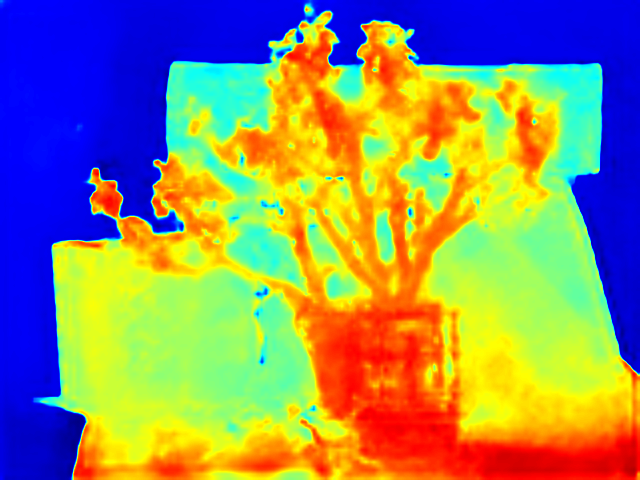}
         &
         \includegraphics[width=3cm]{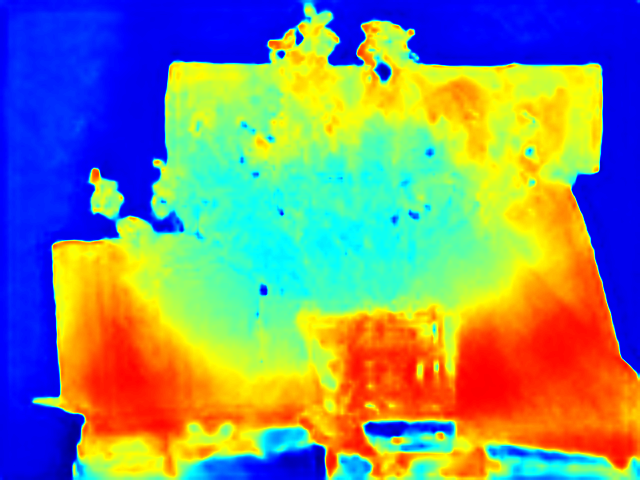}
         \\
         \includegraphics[width=3cm]{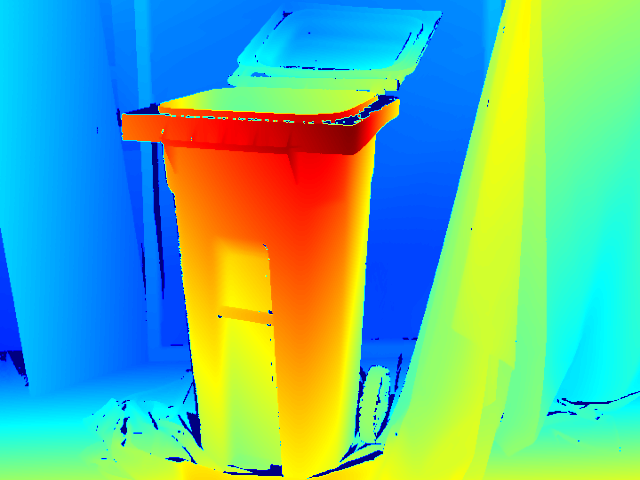}
         &
         \includegraphics[width=3cm]{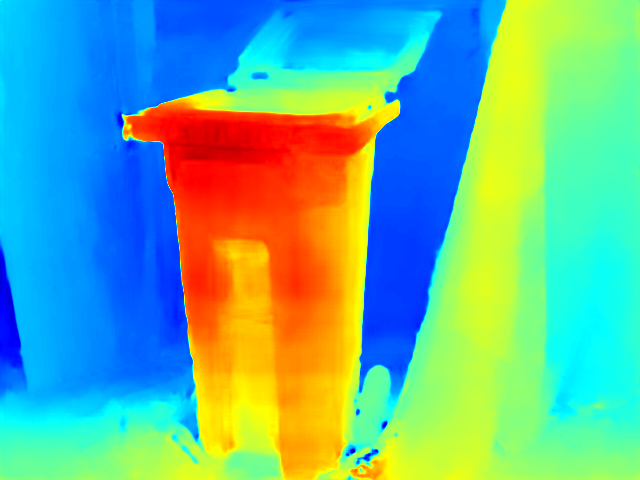}
         & 
         \includegraphics[width=3cm]{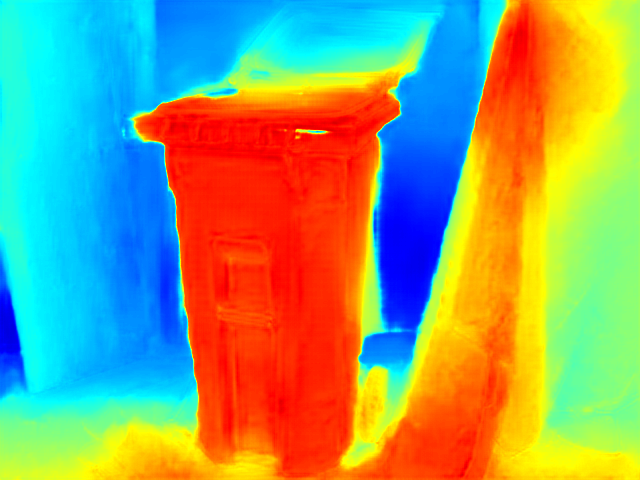}
         &
         \includegraphics[width=3cm]{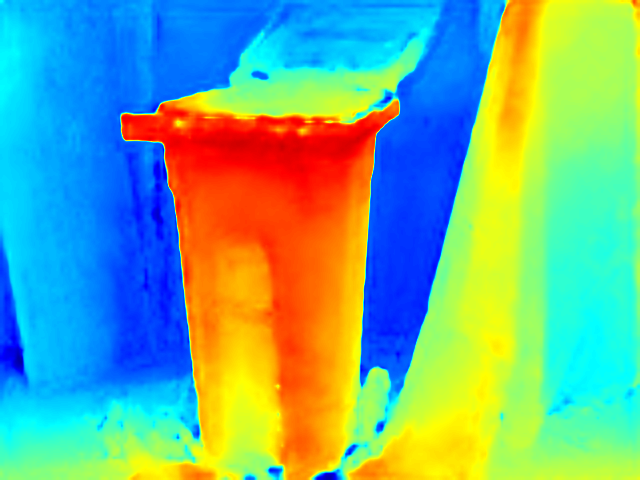}
         &
         \includegraphics[width=3cm]{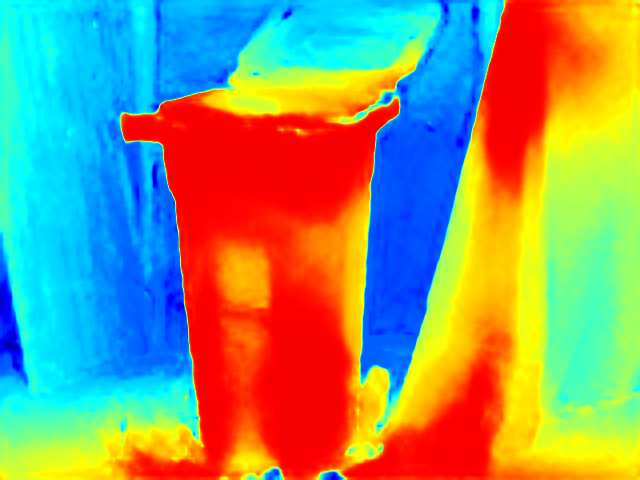}
    \end{tabular}
    }
    \caption{Qualitative results on simulated focal stacks. With the AAT scheme, both network models predict more accurate and finer depth maps, while the non-AAT models fail to distinguish between adjacent objects and also mispredict the depth of some edge objects.}
    \label{fig:supp_simu_comparison}
\end{figure*}

For the experiments on simulated focal stacks, we evaluate the generalizability of the models by training them with focal stacks generated on synthetic RGBD images and testing them with focal stacks generated on real-world RGBD images. There is a significant domain gap between semantic information of the training and testing datasets. However, the experimental results show that the domain gap can be bridged effectively using our proposed AAT scheme. Additional evaluation results are presented in Fig.\ref{fig:supp_simu_comparison}. 

It is worth noting that our results are not contradictory to those in the original papers\cite{wang2021bridging, yang2022deep}, as the training and testing datasets used in those papers were also simulated with the same lenses (thin lens model of the same focal length and aperture), which is consistent with the conclusion we are making.

\bibliographystyle{IEEEtran}
\bibliography{ref}

\ifpeerreview \else
%%%% For the camera ready version, please fill out this
%%%% biography. Your camera ready should be within a 12 page limit
%%%% including acknowledgments, references and biography.

% \includegraphics command within an optional argument. (You could
% create your own custom macro containing the \includegraphics command
% to make things simpler here.)
% \begin{IEEEbiography}[{\includegraphics[width=1in,height=1.25in,clip,keepaspectratio]{mshell}}]{Michael Shell}

% \end{IEEEbiography}

%\newpage


%\vfill

\fi